\documentclass[a4paper, 10pt, conference]{ieeeconf}      
\usepackage{FG2026}

\FGfinalcopy 

\IEEEoverridecommandlockouts                              
\usepackage{graphics} 
\usepackage{epsfig} 
\usepackage{amsmath} 
\usepackage{amssymb}  
\usepackage{microtype}
\usepackage{tikz}
\usepackage{url}

\renewcommand{\paragraph}[1]{\vspace{.5em}\noindent\textbf{#1.}}
\def\FGPaperID{0079} 

\ifFGfinal\else
    \usepackage[pagebackref,breaklinks,allbordercolors=white]{hyperref}
\fi

\usepackage[capitalize]{cleveref}
\crefname{section}{Sec.}{Secs.}
\Crefname{section}{Section}{Sections}
\Crefname{table}{Table}{Tables}
\crefname{table}{Tab.}{Tabs.}

\usepackage{tabularray}
\UseTblrLibrary{booktabs}

\usepackage{siunitx}
\usepackage{pifont} 

\usepackage{bm}

\usepackage{caption}

\usepackage{zebra-goodies}
\usepackage{xspace}

\makeatletter
\let\NAT@parse\undefined
\makeatother
\usepackage[numbers,sort&compress]{natbib}

\makeatletter
\DeclareRobustCommand\onedot{\futurelet\@let@token\@onedot}
\def\@onedot{\ifx\@let@token.\else.\null\fi\xspace}

\def\eg{\emph{e.g}\onedot} 
\def\ie{\emph{i.e}\onedot}

\def\etal{\emph{et al}\onedot}
\makeatother

\newcommand{\argmin}{\mathop{\rm arg~min}\limits}

\NewDocumentCommand\DATASETNAME{}{REACH\xspace}
\NewDocumentCommand\METHODNAME{}{REACH-Net\xspace}

\title{\LARGE \bf
\DATASETNAME{}: Hand Pose Estimation from Room Corners
}

\author{
    Shu Nakamura$^*$\quad
    Ryo Kawahara$^*$\quad
    Genki Kinoshita$^*$\quad
    Ryosuke Hirai$^*$\\
    Yasutomo Kawanishi$^\dagger$\quad
    Shohei Nobuhara$^\ddagger$\quad
    Ko Nishino$^*$\\[0.4em]
    {\tt\small \url{https://vision.ist.i.kyoto-u.ac.jp/}} \\[0.4em]
    $^*$Graduate School of Informatics, Kyoto University \quad
    $^\dagger$RIKEN \quad
    $^\ddagger$Kyoto Institute of Technology \\
}

\newcommand{\minipagepadding}{2mm}
\makeatletter
\let\@oldmaketitle\@maketitle%
\renewcommand{\@maketitle}{
    \@oldmaketitle%

    \begin{center}
        \includegraphics[width=\linewidth]{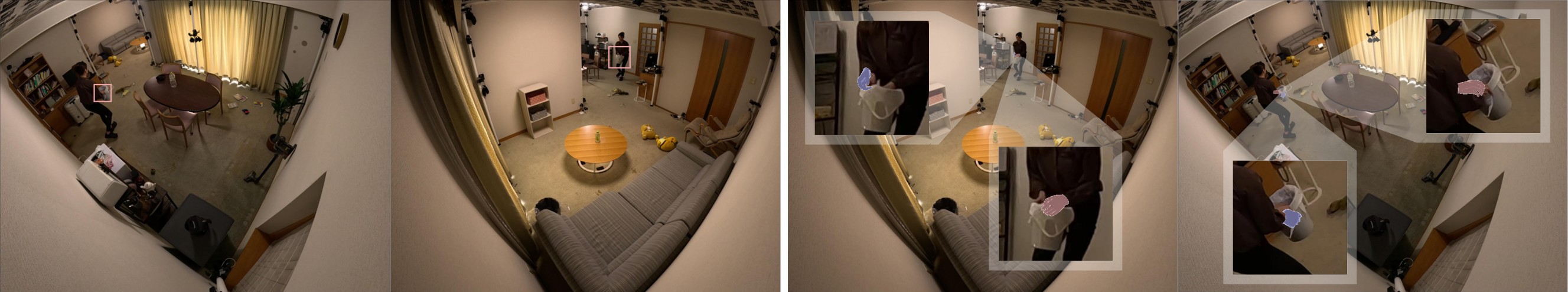}
    \begin{minipage}{0.49 \linewidth}
        \hspace{\minipagepadding}
        \centering
        (a) Two input views
        \hspace{\minipagepadding}
    \end{minipage}
    \begin{minipage}{0.49 \linewidth}
        \hspace{\minipagepadding}
        \centering
        (b) Estimated 3D hand poses of both hands
        \hspace{\minipagepadding}
    \end{minipage}
    \captionof{figure}{
        We achieve accurate hand pose estimation from videos captured from afar, typically from a few (2 or 3) cameras mounted at room corners observing a freely moving person.
        (a) shows two views input to our model. Notice the heavy occlusions and significantly limited resolution of the hands (25 pixels in a 2.7K image). As shown in (b), we fully leverage body coordination and multiview features to successfully estimate hand poses even in such extreme viewing conditions.
    }
    \label{fig:paper_overview}
    \vspace{-1\baselineskip}
    \end{center}
}
\makeatother

\usepackage{fancyhdr}
\begin{document}

\ifFGfinal
\thispagestyle{empty}
\pagestyle{empty}
\else
\author{Anonymous FG2026 submission\\ Paper ID \FGPaperID \\}
\pagestyle{plain}
\fi
\maketitle
\thispagestyle{fancy}

\begin{abstract}

    We introduce a novel 3D hand pose estimator that can accurately recover the shape and pose of people's hands in a room from afar, typically from fixed cameras at room corners, in extremely low-resolution and frequently occluded views. Our key idea is to fully leverage hand-body coordination, its temporal progression, and multiview observations. 
    We achieve this with a novel Transformer-based model, in which hand and body configurations are modeled through correlations between their visual features expressed as per-view tokens, and their temporal coordination is exploited in an autoregressive manner.  
    We introduce a novel dataset which we refer to as \DATASETNAME{}, \textit{R}oom-\textit{E}nvironment dataset \textit{A}nnotated with \textit{C}hest cameras for \textit{H}and pose estimation, to train and test our method.
    \DATASETNAME{} is a first-of-its-kind large-scale hand pose dataset that captures accurate hand movements of 50~participants across a wide variety of daily activities. In order to avoid interfering with natural movements while annotating the hands with accurate shape and pose, we leverage concealed chest cameras. 
    Through extensive experiments, including comparative studies with existing methods, we show that our model, \DATASETNAME{}-Net, achieves highly accurate 3D hand pose estimation from afar. These results broaden the horizon of 3D hand pose estimation, especially towards ``in-the-wild'' continuous human behavior analysis. 

\end{abstract}

\section{Introduction}
3D hand pose estimation plays a critical role in understanding human actions and their underlying intents.
Continuously estimating the hand pose of freely moving people without obstructing their bodies or sight can help better understand people's natural activities in their daily lives. Invasive hand pose capture using wearable cameras, markers, and devices (\eg, gloves and wristbands) may be acceptable for certain applications, such as VR experiences, but becomes a significant burden for daily use. We envision using 3D hand poses as visual cues for understanding people's behavior in a room, for instance, monitoring the safety of the elderly, looking after children, and gauging their inner states. The ability to accurately estimate the hand pose in such daily scenes can enrich the clues to decipher the intent and mood of people, \eg, visually understanding that a person is trying to pass a cup to another one rather than using it herself, which can have broad implications for human or robotic intervention. For this, non-invasive capture and estimation become critical. 

Past works on visual hand pose estimation, as we will review in detail later, require ego-views~\cite{UmeTrack2022}, typically using glass-mounted cameras, or inherently rely on close exo-views captured from relatively short distances~\cite{epflSmartKitchen30_2025}, typically up to \qty{4}{m}. These approaches fundamentally constrain either the person, as they need to wear a camera, or the scene, as many cameras need to be placed in it, which becomes apparent not just to the person but also in any view capturing the person. Ideally, we would like to accurately track the 3D hand pose using a handful of cameras mounted at room corners, \eg, on the ceiling, so that they do not obstruct the person's activities or clutter the scene. 

The key challenge in achieving accurate 3D hand pose estimation from such fixed-viewpoint cameras is that the viewpoints are far from the hands, typically \num{5} to \qty{10}{m}, resulting in very low resolution of the hand (\eg, \num{20} px in a 1080p feed) for a regular field of view. 
Even worse, as the viewpoints are static and do not move with the hands, the hands are frequently occluded by the room itself, by the person herself, and also by the objects she is holding, resulting in even zero hand-region pixels in a given view. 
Most past works that rely on single-view observations of hands~\cite{hamer,yu2025dynhamr,hmp,wilor2024} inevitably fail, as the hands are too small or completely occluded.
Even when multiple views are available, it is not apparent how to leverage them, as there is no clear way to tell which camera has a clearer shot of the hands.

How can we achieve 3D hand pose estimation from afar? How can we obtain accurate 3D hand poses of people in a room with cameras simply mounted at room corners? 
Our key insight is that non-closeup views capture the whole body of the person, giving us the opportunity to fully leverage the strong contextual cues carried by the whole-body posture and movements to pin down the 3D hand pose from sparse observations, both in resolution and viewpoints. Certain body postures can indicate a particular hand pose, which could help resolve the lack of fidelity in hand appearance. 
Leveraging body pose, however, is not trivial, as the relationship between the subtle movements of the hands and the overall body posture is highly complex and nonlinear.
Exploiting multiview cameras is also not straightforward in the absence of a means to know which camera view provides the most useful information at any given moment (we would need to know the hand pose to know that!). 
Our key contribution is a novel hand pose estimator that fully exploits the hand body coordination to achieve accurate hand pose estimation from afar while being robust to frequent occlusions. We refer to this novel hand pose estimator as \METHODNAME. \METHODNAME integrates visual features of the hands, the body, and body posture from different views in an encoder-decoder Transformer architecture, which enables full capitalization of multiple far views. \METHODNAME fully exploits the temporal coordination of the hands and body across these multiple views in an autoregressive manner. 

We need a dataset that captures a variety of natural human activities in a real room environment with sparse but multiple far-view cameras to train and evaluate our proposed method.
To the best of our knowledge, such a dataset does not exist, and all existing datasets either capture people with head-worn cameras and markers or clutter the scene with near-distance cameras scattered throughout the environment, which inevitably biases the appearance of both the people and the scenes. 
We introduce a novel dataset, which we refer to as \textbf{R}oom-\textbf{E}nvironment dataset \textbf{A}nnotated with \textbf{C}hest cameras for \textbf{H}and pose estimation (\DATASETNAME{}).
\DATASETNAME{} captures more than \num{540000} frames of \num{50} people carrying out a wide ranee of activities involving various types of objects in a real living room.
Accurate hand pose annotation is achieved using small concealed chest-mounted cameras and markers on the ceiling, both of which are hardly visible from the far-view cameras. \DATASETNAME{} lets us train our model and thoroughly evaluate its accuracy in fair settings against existing methods. 
We show that our novel method achieves higher precision compared with past methods, especially in challenging situations with severe occlusions. 
Our code and dataset are available on our project page. 

Our contributions in this paper are threefold.

\begin{enumerate}
    \item \METHODNAME{}, a novel method for 3D hand pose estimation from far views at extremely low resolution and under heavy occlusion.
    \item \DATASETNAME{}, a large-scale dataset of ``in-the-wild'' indoor activities of people with accurate 3D hand poses.
    \item Extensive experiments demonstrating the effectiveness of \METHODNAME{} and its state-of-the-art accuracy for challenging 3D hand pose estimation from afar.
\end{enumerate}

\section{Related Works}
We start by reviewing past methods and datasets on hand pose estimation.

\subsection{Hand Pose Dataset}
Dataset collection for hand pose estimation inherently suffers from a chicken-and-egg problem: to collect a large-scale dataset automatically, an accurate hand-pose estimation method would become critical, but to train such a method, we need a large-scale dataset.
To address this dilemma, past works measure hand pose using different modalities, such as measurement gloves~\cite{parahome2024}, magnetic sensors~\cite{egoHandPoseEst-Hernando2018}, head-mounted cameras~\cite{fan2023arctic,ego-exo4d,hoi4d,hot3d_2024_banerjee,epflSmartKitchen30_2025}, and motion capture~\cite{grab2020, trumans}.
The drawback of these methods is that they require special devices to be worn by the participants, which inevitably alters the appearance and hinders the natural movements of the participants.
To alleviate this problem, some works use one or more RGB-D cameras~\cite{HO3D_honnotate, egoHandPoseEst-Hernando2018, ContactPose,intercap2022, DexYCB, hoi4d} while other works employ densely installed RGB cameras~\cite{Assembly101,FreiHAND,PanopticHandKeypointMVBootstrapping2017} and exploit multiview geometry to recover the hand poses.
These methods, however, require closeup camera views, which often restrict the movements of the participants and visually clutter the environment causing inherent bias in the visual data. 

IKEA-ASM~\cite{IKEA-ASM2021} collects natural assembly actions in a room-scale environment from a relatively far viewpoint (\qty{3}{m} to \qty{4}{m}), but since the hand pose is not annotated, it is not suitable for training hand pose estimation models.
Panoptic Hand Keypoint Dataset~\cite{PanopticHandKeypointMVBootstrapping2017} annotates hand keypoints in the world coordinate system using a bootstrap method that iteratively repeats the training of the keypoint estimator and the annotation of the hand keypoints by triangulating the keypoints estimated using the model.
This method works well when a large number of viewpoints are available (31 cameras are used in their experiment) and the quality of the dataset degrades as the number of viewpoints goes down.
When participants freely move in a room-scale environment, this assumption is not satisfied as it is unrealistic to install such a large number of cameras in regular houses.

Other datasets such as HInt~\cite{hamer} and WHIM~\cite{wilor2024} collect hand images from existing datasets or web-crawled videos, but these datasets lack multiview observations, which are crucial for combating occlusions. 
The closest to our desiderata is RICH~\cite{rich_bstro_2022}, which annotates SMPL-X poses through penalizing intersections with the scene mesh. The accuracy of hand pose estimation is, however, not guaranteed, as it is estimated from images taken from afar and annotation errors are not evaluated.

In summary, as shown in \cref{tab:dataset_comparison}, existing datasets do not capture hand poses in the world coordinate system in natural-looking (\ie, visually uncluttered daily scene) room-scale environments with object interactions. These, together with multiview observation, are essential for deriving and evaluating methods for hand pose estimation from afar in practical environments.

\begin{table*}[t]
    \newcommand{\cmark}{\textcolor{green}{\ding{51}}} 
    \newcommand{\xmark}{\textcolor{red}{\ding{55}}} 
    \newcommand{\noVisibleDevices}{No Vis.\ Dev.}
    \newcommand{\naturalImage}{Nat.\ Img.}
    \newcommand{\multiViewHandPoseAnnotation}{M/V HPA}
    \newcommand{\roomScaleArea}{Room-Scale}
    \newcommand{\handActionLabels}{Act.\ Labels}
    \newcommand{\Modalities}{Modalities}
    \centering
    \SetTblrInner{rowsep=2pt,colsep=6pt,stretch=1,abovesep=2pt}
    \begin{tblr}{@{}lcccccccc@{}}
        \toprule
        Dataset                                  & \noVisibleDevices & \naturalImage & \multiViewHandPoseAnnotation & \roomScaleArea & \handActionLabels & \Modalities & \# of subj. & \# of Frames \\
        \midrule
        MuViHand~\cite{muvihand}                 & \cmark            & \xmark        & \cmark                       & \xmark         & N/A               & RGB         & N/A         & 402K         \\
        TRUMANS~\cite{trumans}                   & \cmark            & \xmark        & \cmark                       & \cmark         & Provided          & RGBD        & N/A         & 1.6M         \\ 
        MVHM~\cite{mvhm2021}                     & \cmark            & \xmark        & \xmark                       & \xmark         & N/A               & RGB         & N/A         & 320K         \\
        ParaHome~\cite{parahome2024}             & \xmark            & \xmark        & \cmark                       & \cmark         & H/O contact       & N/A         & 38          & 875K         \\
        DexYCB~\cite{DexYCB}                     & \cmark            & \cmark        & \cmark                       & \xmark         & N/A               & RGBD        & 10          & 582K         \\
        Ego-Exo4D~\cite{ego-exo4d}               & \cmark            & \cmark        & \xmark                       & \cmark         & Provided          & RGB         & 740         & 139M         \\
        ARCTIC~\cite{fan2023arctic}              & \xmark            & \cmark        & \cmark                       & \xmark         & H/O Contact       & RGB         & 10          & 243K         \\
        HOI4D~\cite{hoi4d}                       & \xmark            & \cmark        & \xmark                       & \cmark         & Provided          & RGBD        & 4           & 2.4M         \\
        IKEA-ASM~\cite{IKEA-ASM2021}             & \cmark            & \cmark        & \xmark                       & \cmark         & Provided          & RGB         & 48          & 3M           \\
        HInt~\cite{hamer}                        & \cmark            & \cmark        & \xmark                       & \xmark         & N/A               & RGB         & N/A         & 40.5K        \\ 
        WHIM~\cite{wilor2024}                    & \cmark            & \cmark        & \xmark                       & \xmark         & N/A               & RGB         & N/A         & 2M           \\
        InterCap~\cite{intercap2022}             & \cmark            & \cmark        & \cmark                       & \xmark         & Use/Check/Pass    & RGBD        & 10          & 67K          \\
        EPFL-SK30~\cite{epflSmartKitchen30_2025} & \xmark            & \cmark        & \cmark                       & \cmark         & Provided          & RGBD        & 16          & 321K         \\ 
        \textbf{REACH (Ours)}                    & \cmark            & \cmark        & \cmark                       & \cmark         & N/A               & RGB         & 50          & 540K         \\
        \bottomrule
    \end{tblr}
    \caption{Comparison of different human hand pose datasets. Each column describes the following.
        \noVisibleDevices{}: whether the dataset is captured without any visible devices;
        \naturalImage{}: whether the dataset comes with natural images;
        \multiViewHandPoseAnnotation{}: whether the hand pose is annotated from more than one view;
        \roomScaleArea{}: whether the dataset is captured in a room-scale area;
        \handActionLabels{}: whether the dataset provides action labels such as hand-object contact; and
        \Modalities{}: the image modalities provided by the dataset.
    \DATASETNAME{} is the first large-scale dataset that captures hand pose in a room-scale environment from multiple viewpoints with natural appearance and movement.}
    \label{tab:dataset_comparison}
    \vspace{-1\baselineskip}
\end{table*}

\subsection{Monocular Hand Pose Estimation}
With the advent of VR and AR headsets, single-view hand pose estimation has become a popular research topic. 
HaMeR~\cite{hamer} is a ViT~\cite{ViT}-based hand mesh estimation model that directly estimates MANO~\cite{mano2017} parameters from monocular images.
WiLoR~\cite{wilor2024} is a state-of-the-art monocular hand pose estimation model that obtains an initial estimate using a ViT-based Transformer and further refines it using an output feature map.
Fan \etal~\cite{fan2025_robust_lowres_hand_rec} propose a hand pose estimation method that leverages temporal hand and finger joint features to construct robust low-resolution features. This method, however, is designed only for ``slightly distant'' scenes (up to a few meters), where fingers are still well discernible. In contrast, our focus is on room-scale environments where cameras do not obstruct the scene itself, \ie, located at room corners. In this case, the distance is more than a few meters, typically ranging from \num{5} to \qty{10}{m} and the hand and finger joints are not discernible. A direct comparison with our method is, however, impossible as Fan \etal do not provide the trained model and code. Our code and dataset are available on our project page. 

To achieve hand pose estimation at this distance, we leverage hand-body coordination rather than just looking at the hand.
Some methods leverage body pose as a prior for hand pose estimation~\cite{body2hands2021, body2HandTransformer2023, liu2022spatial} by first estimating the body pose and then estimating the hand pose from it. These methods are restricted to single views. 
Such cascaded modeling also cannot fully leverage the complex integrated coordination of the hand and body. We tap into this strong contextual cue with a novel Transformer architecture and leverage temporal coordination through autoregressive estimation. 

\subsection{Multiview Hand Pose Estimation}
A few methods have been proposed for leveraging multiview observations of the hands. 
MuViHand-Net~\cite{muvihand} is an LSTM-based hand pose estimation method that leverages multiview video input, but it requires a fixed number of observations, fixed viewing directions, and closeup tracking of the hands. UmeTrack~\cite{UmeTrack2022} takes images from different viewpoints as an optional input and adopts a recurrent network to robustly estimate temporally consistent hand pose to combat occlusions.
MVHM~\cite{mvhm2021} also proposes a multiview hand pose estimator together with a multiview dataset. 
These methods, however, assume that high resolution hand images are available as input. Our focus is on hand pose estimation from a far distance for room-scale non-invasive monitoring, where closeup views are not available.

\section{\DATASETNAME{} Dataset}
\begin{figure*}[ht]
    \centering
    \includegraphics[width=\linewidth]{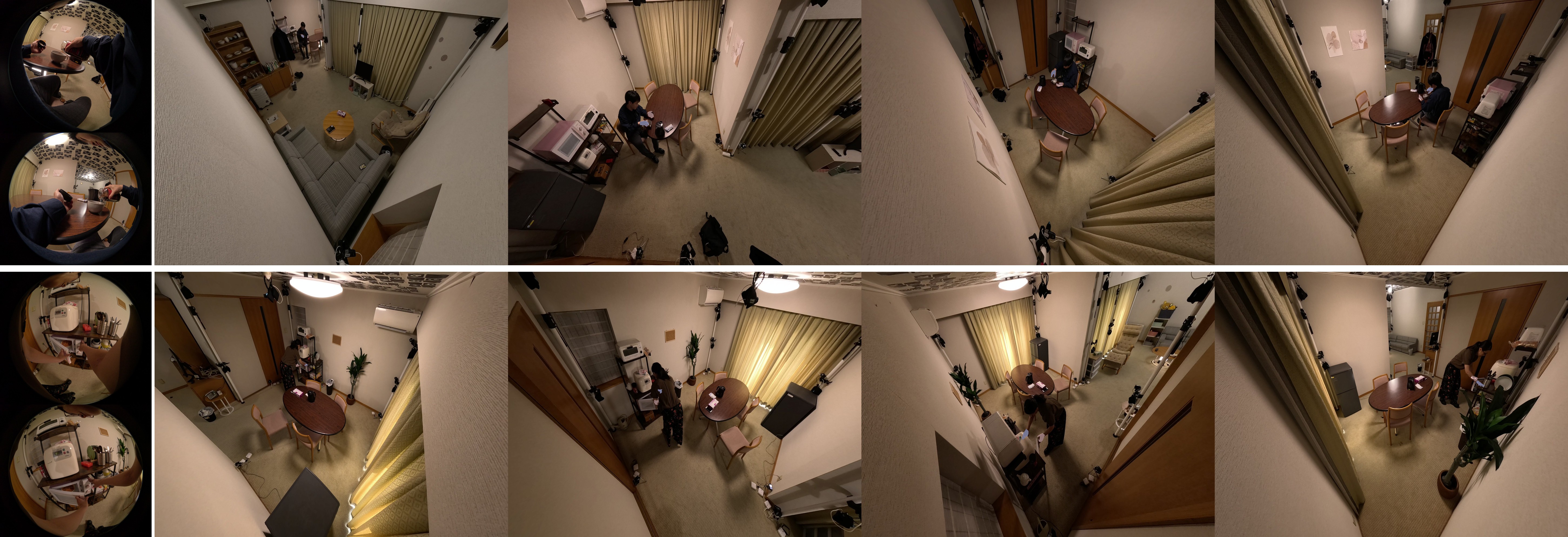}
    \caption{Sample images from the \DATASETNAME{} dataset.
    The leftmost column shows images captured by the chest cameras (used only for annotation), and the other columns show images captured by the fixed-view room cameras. 
    The ground truth hand (red for the left hand and blue for the right) is overlaid on the fixed-view images. The hand pose is estimated from the chest cameras and triangulated to the world coordinate system.}
    \label{fig:dataset_examples}
    \vspace{-1\baselineskip}
\end{figure*}

We first introduce our novel dataset \DATASETNAME{}, which captures natural daily human actions from wall-mounted cameras with accurate 3D hand pose and position annotations in the world coordinate system. \DATASETNAME{} is the first of its kind in scale, natural appearance, non-invasiveness, and annotation. 
As shown in \cref{fig:dataset_examples}, hand poses are annotated in the world coordinate system and can be projected to any camera easily.
One underlying key idea to achieve this large-scale annotation is to capture closeup hand images from chest cameras worn by the participants which are concealed from the fixed far-view cameras.
This data acquisition was approved by the Institutional Review Board at Kyoto University (KUIS-EAR-2020-002). Informed consent was obtained from all participants.

\subsection{Dataset Characteristics}
To train and evaluate a model that can estimate hand pose from multiview RGB images or video input from far viewpoints, the dataset must meet two requirements.
First, it needs to be captured in an environment with a natural appearance, with no markers or measurement devices visible to fixed cameras.
Second, the environment should be large enough and capture interactions with various types of household objects to cover a wide range of activities that may take place in daily indoor scenes.
As shown in \cref{tab:dataset_comparison}, to the best of our knowledge, \DATASETNAME{} is the first dataset that fulfills these requirements.

The dataset is captured in a real living room of about \qty{60}{m^2} with various furniture and objects, including a wardrobe, a TV monitor, a dining table, a remote for an air conditioner, stuffed animals, pens and papers, desk lights, and a fridge.
In this room, we captured sequences of 50 participants (27 males and 23 females) whose ages are evenly distributed between their twenties and sixties.
We asked each participant to engage in various daily activities, such as ``wipe the table,'' ``take books from the shelf and read them on the sofa,'' ``take a remote control and change the channel,'' and ``fold the clothes on the floor and put them in the wardrobe.''

The room is covered by 16 cameras installed in the ceiling corners and on the walls.
Typically 2 to 4 of them are used for training and inference.
The cameras are synchronized before and after the capture using a strobe light, and they capture videos at \qty{60}{FPS} with a resolution of \qtyproduct{2704 x 2028}{px}.
Since the cameras are in the corners of the room, the participants' natural movements are not hindered.
The camera poses and ceiling marker positions are calibrated before the capture.
In total, we capture 1M frames of videos from the fixed cameras, and 540K frames of the videos capture the hand from the chest cameras.

\subsection{Automatic Data Annotation}
We annotate, for every single frame, the MANO~\cite{mano2017} hand parameters (shape and joint angle parameters) and the hand position in the world coordinate system.
Annotating them is hard since the hand is often occluded by various objects in the environment and is often very far from the fixed camera, exactly the challenges we would like to overcome in this work to make hand pose estimation practical in the wild.

To avoid these problems, we use chest-mounted fisheye cameras and completely conceal them with a jacket so that they do not bias the captured appearance of the person. 
We also installed ArUco markers~\cite{aruco} on the room ceiling to accurately estimate the chest camera poses in the world coordinate system. Note that these markers are not located around the participants or scene objects, so they effectively do not affect the appearance of the captured room. This again is important to not bias the scene appearance. 

\begin{figure}[t]
    \centering
    \includegraphics[width=0.49 \linewidth]{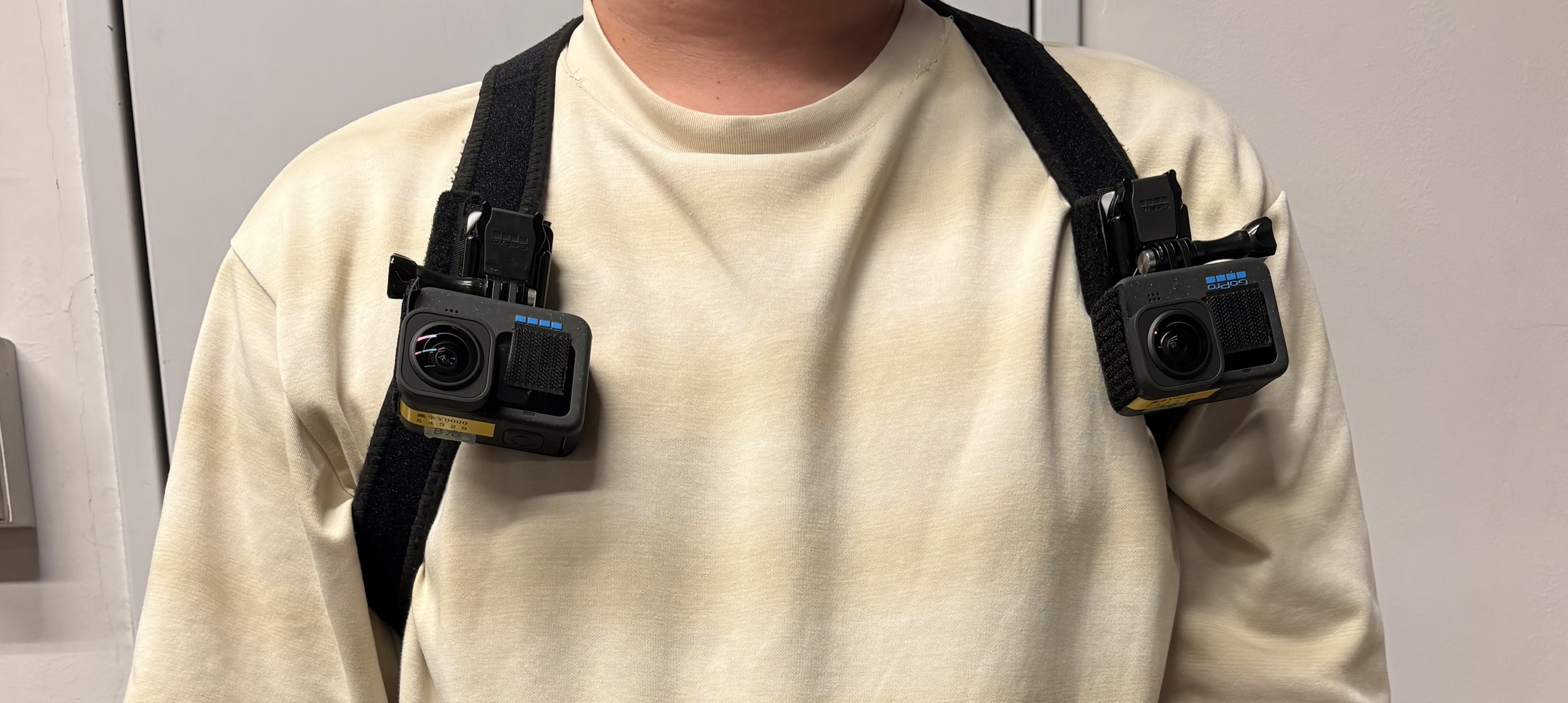}
    \includegraphics[width=0.49 \linewidth]{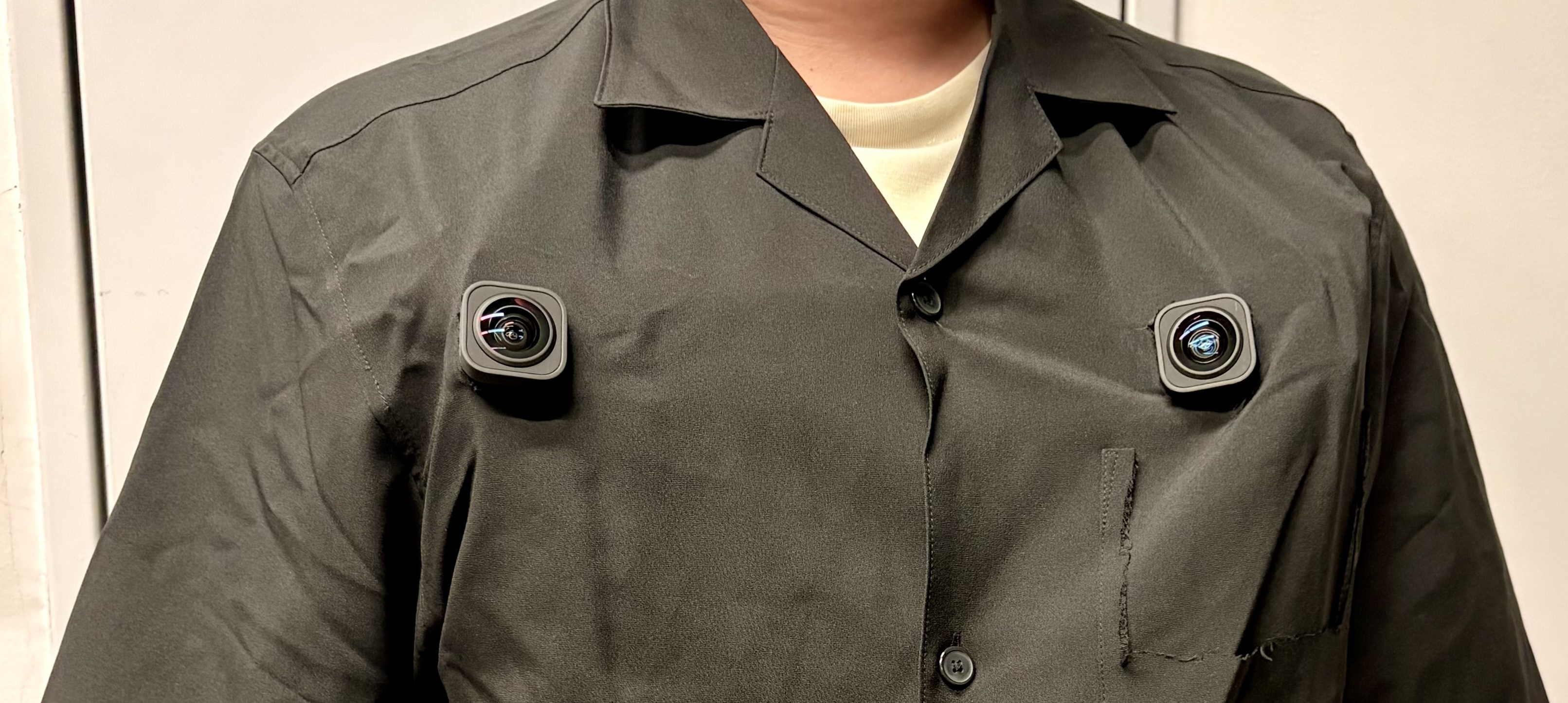}
    \caption{Clothing for the capture. During capture, the participants wore a chest strap mounted with two fisheye cameras and a jacket over it. This setup (a) captures closeup views of both hands with two fisheye cameras, (b) does not alter the natural appearance of the person, and (c) does not hinder the dynamic movements occurring in daily activities.}
    \vspace{-1\baselineskip}
    \label{fig:clothes}
\end{figure}

As shown in \cref{fig:clothes}, we mount two GoPro 12 cameras on the participants' chests using a chest strap.
The GoPro cameras are equipped with the Max Lens Mod 2, which has a \ang{175} field of view.
The chest cameras capture the hands as long as they are in front of the chest, which is the case during most daily activities. Note that, unlike head-worn cameras (\eg, VR/AR headsets), the person does not need to be looking at her hands, which significantly broadens the daily activities we can capture in the dataset, well beyond those that naturally require focusing on one's own hands like cooking. 
We ask participants to wear a jacket over the cameras, which conceals them so they are not visible to the room cameras. 
The jackets have two holes on the chest where the cameras are mounted and whose size is designed to be just large enough for the camera lenses to peek through.
The chest cameras and markers on the ceiling are hardly visible and the environment appears perfectly natural as can be seen in \cref{fig:paper_overview,fig:dataset_examples}.

The wide field of view of the chest cameras enables them to capture the ceiling ArUco markers which are used to estimate the camera pose in the world coordinate system.
We combine two methods for camera pose estimation: PnP with the ceiling markers and ORB-SLAM3~\cite{orb-slam3}.
The dense camera trajectory estimated by ORB-SLAM3 is transformed into the world coordinate system using the marker-estimated camera poses. 

Obtaining the hand pose from chest camera videos is not straightforward, as the hand can be strongly distorted by the fisheye lens.
We adopt a three-step approach that first detects the approximate hand region, then estimates the hand mesh from the undistorted image of that region, and finally computes the original hand orientation in the world coordinate system using the estimated chest camera pose.
We detect the hand region in the fisheye image using MediaPipe~\cite{mediapipe}, which provides the bounding boxes of the hand in the chest camera images.
Given the bounding box, we obtain a virtual camera that is undistorted, targeted at the center of the bounding box, and has the same FoV as the longer side of the bounding box.
We denote the rotation of this virtual camera relative to the chest camera as $\bm{R}^\text{C}_i$.
We then resize the cropped image to \qtyproduct{256 x 256}{px} and use HaMeR~\cite{hamer} to estimate the MANO shape parameter $\beta$, pose parameter $\theta$, and hand orientation $O$.
The hand orientation in the world coordinate system is calculated by converting the orientation $O^\text{W} = {\left(\bm{R}\bm{R}^\text{S}_i\right)}^\top{\bm{R}^\text{C}_i}^\top O\,$.
Please see supplementary material for details to obtain the rotation to the world coordinate system $\bm{R}\bm{R}^\text{S}_i$.
The hand mesh is then scaled to match the pre-acquired physical hand sizes for each subject.

To annotate the hand locations in the world coordinate system, we use the triangulated wrist keypoints from the fixed cameras as we found that the estimated translations of the chest cameras are not accurate enough.
Please see supplementary material for details.

\Cref{fig:dataset_examples} shows sample frames from the \DATASETNAME{} dataset.
The leftmost column shows the example images from the chest cameras. Thanks to our multi-stage processing pipeline, the hand poses are estimated even when they are far from the image center and highly distorted by the fisheye lens.
The other columns show example images from the fixed cameras, overlaid with the ground-truth hand pose.

\subsection{Limitation}
Chest-mounted cameras cannot capture hand poses when the hands are behind the body or too low for the chest cameras to see. Note, however, that most daily activities involving hand movements occur in front of the body and at a reasonable height, which the wide FoV captures, as shown in the dataset.
The chest-mounted cameras may also hinder some upper-body movements, but we found that the participants could carry out most daily activities naturally and there was no report from participants that they could not do certain actions due to the cameras.

\section{\METHODNAME{}}
\begin{figure*}[ht]
    \centering
    \includegraphics[width=\linewidth]{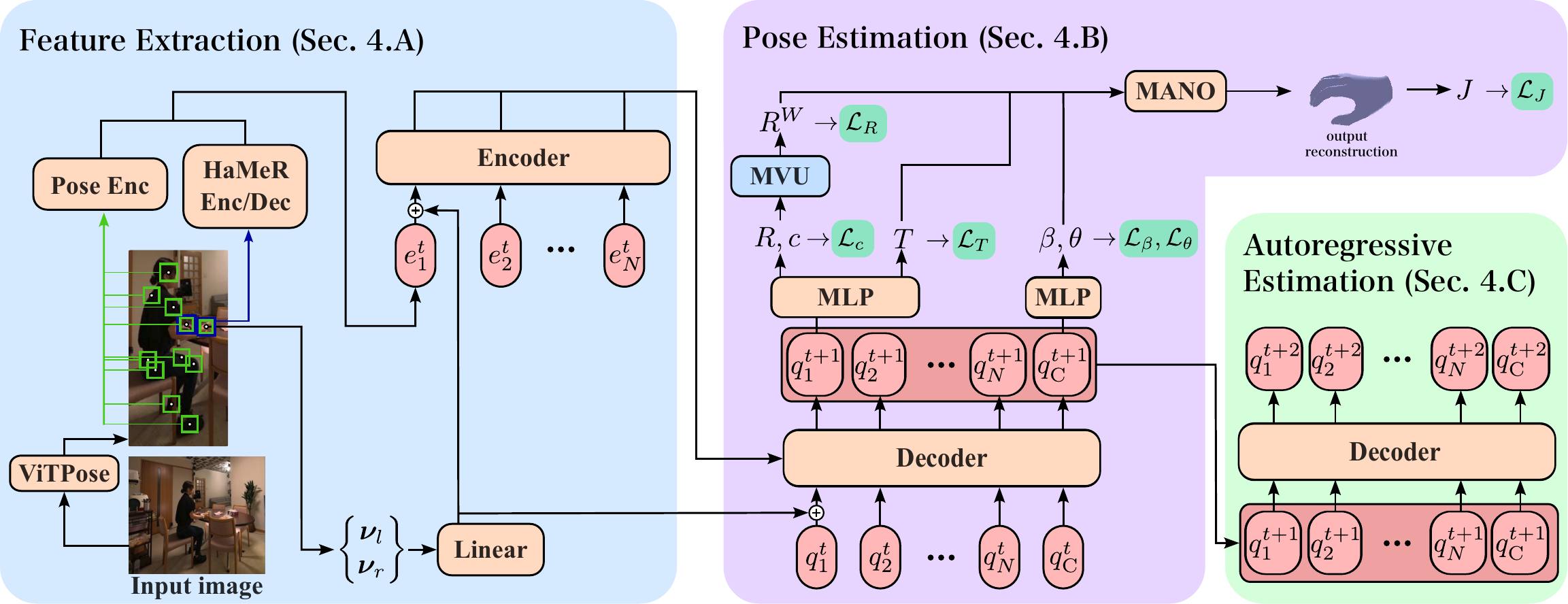}
    \caption{Network architecture of \METHODNAME{}. The input is multiview videos, and the output is a 3D hand reconstruction. After extracting features from each frame using ViTPose~\cite{vitpose} to produce tokens $e^t_n$, we process them using a Transformer encoder-decoder architecture.  The decoder's output tokens are used to estimate hand pose parameters. They are then used to render the hand joints in each camera's coordinate system.}

    \vspace{-1\baselineskip}
    \label{fig:_images_method_oneframe}
\end{figure*}

We derive \METHODNAME{}, a novel 3D hand pose estimator which autoregressively estimates 3D hand pose from far, multiview video. 
\METHODNAME{} fully leverages hand-body coordination to accurately and robustly estimate hand poses from afar.
As shown in \cref{fig:_images_method_oneframe}, we carefully design \METHODNAME{} as a Transformer-based autoregressive model that takes multiple far-view videos as input, and outputs 3D hand pose estimates for each viewpoint. The method only uses the relative camera poses, which is essential for generalizing to arbitrary, calibrated camera setups in real environments. 
Specifically, the input is videos from a few (typically 2 or 3) cameras installed on the ceiling and walls of a room, and the output is the MANO parameters for each hand $\{\beta_c, \theta_c\} _{c \in \{l, r\}}$ and their rotations and translations from each view $\{R_{c,n}, T_{c,n}\} _{c \in \{l, r\}, n \in 1\ldots N}$.

\subsection{Hand and Body Coordination}

\NewDocumentCommand\HANDFEATURES{}{Hand Features}
\NewDocumentCommand\BODYFEATURES{}{Body Features}
\NewDocumentCommand\PoseEncoder{}{Pose Encoder}

For each input image $I_{n,t}$ from each viewpoint $n \in \{ 1\ldots N\}$ at time $t \in 1\ldots T$, we extract \HANDFEATURES{} and \BODYFEATURES{}.
\HANDFEATURES{} are obtained from cropped hand images to leverage hand appearance when not occluded.
\BODYFEATURES{} combine global image features and body joint locations to encode the whole-body pose and appearance, guiding robust hand pose estimation even during occlusion.

A square region is cropped around each hand, with width and height set to half the target person's height computed from the keypoints, and the resulting region is passed to the Transformer backbone of HaMeR~\cite{hamer} to obtain the hand pose features.
To compute \BODYFEATURES{}, we extract \num{256}-dimensional visual features of the input images from the third block of ImageNet~\cite{imagenet}-pretrained ResNet~\cite{resnet2016he} and ROI align~\cite{mask-rcnn} them around 2D body keypoints estimated by ViTPose~\cite{vitpose}.
Two types of learnable positional encodings are added to describe the poses of the body joints. The first is the one from relative joint positions to the center of the person's bounding box, and the second is the indices of the body joints. These values are linearly projected to the size of the visual features.
To model the relationships among body joints, we process these features with a lightweight Transformer encoder, which we refer to as \PoseEncoder{}. We obtain the \BODYFEATURES{} from its CLS token output, which is initialized with a learnable vector.
\HANDFEATURES{} and \BODYFEATURES{} are then concatenated to obtain the image feature $\left\{e_n^t\right\}_{n \in 1\ldots N}$.

\subsection{Multiview Coordination}

\METHODNAME leverages the Transformer encoder-decoder architecture to merge intra- and inter-image features for robust hand pose estimation across arbitrary views.

The decoder at time $t$ is provided with the query tokens $\{q_n^t\}_{n \in 1 \ldots N}$ and the CLS token $q^t_\text{C}$.
Each $q^t_n$ is used for estimating the hand locations (translation $T_n$ and rotation $R_n$) in the cropped hand images from the $n$-th camera, while the CLS token $q^t_\text{C}$ is responsible for estimating the hand posture (hand joint angles $\theta$ and hand shape $\beta$ parameterized by MANO~\cite{mano2017}).
Each camera token is initialized with positional embeddings of view directions, allowing the network to leverage their relationships.
Specifically, for each camera and each time frame the 3D ray direction toward the left and right hand, denoted as $\{\bm\nu_{n,l}, \bm\nu_{n,r}\}$ are computed.
We rotate the world and cancel the first camera's rotation, so that the entire network becomes invariant to absolute camera poses, \ie, only the relative rotations between cameras are relevant. This makes \METHODNAME{} applicable to any camera setup and environment without retraining.
The CLS token is initialized with a learnable fixed vector.
The positional encoding generated from the ray directions of camera $n$ is also added to the image features $e_n^t$ to provide the camera pose information and correspondence to the encoder.

For each of the outputs of the decoder $\{q^t_n\}_{n \in 1 \ldots N}$ and $q^t_\text{C}$, an MLP converts it into the hand pose parameters. The output tokens corresponding to the camera tokens $q^{t+1}_n$ are converted to the rotation and translation parameters from the cameras, $R_n$ and $T_n$ respectively, while the output token corresponding to the CLS token $q^{t+1}_\text{C}$ is converted to the hand joint angles and shape parameters $\beta,\theta$.

The hand rotation parameter $R_n$ cannot be estimated when the hand is occluded in the given view.
To this end, we introduce Multiview Unification (MVU) to consolidate view-dependent hand rotation estimation.
Specifically, we modify the MLP to additionally output rotation confidence scores $c_n \in [0, 1]$ from $q_n$, and take the weighted average of orientation estimation
\begin{equation}
    \tilde{O}^W = G^{-1}\left(\frac{\sum_{n=1}^{N} c_n G\left({\bm{R}^\text{F}_n}^\top O_n \right)}{\sum_{n=1}^{N} c_n}\right)
\end{equation}
where $\bm{R}^\text{F}_n$ denotes the camera rotation parameter.
Here, we denote the conversion from the rotation matrix to 6D rotation representation~\cite{continuous_rot_rep_2019} as $G$ and its inverse as $G^{-1}$.
MVU not only enables robust and consistent hand rotation estimation but also stabilizes training.

\subsection{Temporal Coordination}
For hand pose estimation, in addition to hand-body coordination, temporal changes provide rich contextual cues. 
\METHODNAME{} leverages this by autoregressively estimating the hand pose from the input videos. This can be achieved by initializing the query tokens $q^{t+1}$ with the previous frame's output.
In this way, the model's task is to update the previous frame's hand pose estimate using the current frame's observation. This autoregression is essential for robust and accurate estimation across temporary occlusions as we demonstrate.

\subsection{Training}
The loss function is
\begin{equation}
    \mathcal{L}
    = \lambda_{R}  \mathcal{L}_{R} 
    + \lambda_{T}  \mathcal{L}_{T} 
    + \lambda_{\beta}  \mathcal{L}_{\beta} 
    + \lambda_{\theta} \mathcal{L}_{\theta} 
    + \lambda_{J}  \mathcal{L}_{J}
    + \lambda_\text{C}  \mathcal{L}_\text{C} 
\end{equation}
where $\mathcal{L}_{R}$, $\mathcal{L}_{T}$, $\mathcal{L}_{\beta }$, $\mathcal{L}_{\theta }$, and $\mathcal{L}_{J}$ are the losses for the hand rotation, translation, hand shape parameters, hand joint angles, and rendered hand joints projected onto the image plane, respectively.
We use the L2 loss for $\mathcal{L}_{\beta}, \mathcal{L}_{\theta}$, and $\mathcal{L}_{J}$, and the L1 loss for $\mathcal{L}_{T}$.
$\mathcal{L}_\text{C}$ is a regularization term for the confidence score $c$ to guide the model to have high confidence when the hand is clearly visible. We use the sigmoid cross-entropy of the confidence of the wrist keypoint estimated by ViTPose for $\mathcal{L}_\text{C}$.
We set the number of cameras $N$ to \num{4} for the first stage of training and reduce it to \num{2} for the second stage.
The first stage stabilizes the initial training as the model is served with input videos from a larger number of cameras.

\subsection{Ethical Considerations}
The proposed method is designed for monitoring human activities.
Our method works as an alternative to surveillance systems that typically require a high-resolution video.
It enables the extraction of essential behavioral information (hand pose) from low-resolution images, which makes individual identification difficult. It also works without a human being involved. Thus, we believe our method allows us to address the needs of human behavior monitoring while minimizing the risk of identity leakage.

\section{Experimental Results}
We evaluate the effectiveness of \METHODNAME{} with a set of experiments on \DATASETNAME{} and existing datasets.
We first evaluate the ground-truth annotation accuracy of the \DATASETNAME{} dataset and then compare \METHODNAME{}'s accuracy with several baseline methods. We also conduct ablation studies to validate the effectiveness of each component of \METHODNAME{}.

\subsection{Accuracy of Hand Pose Annotation}
We randomly selected 100 frames from the dataset and manually annotated 2D hand keypoints for all clear views.
We then compare the distance between our ground truth data and the 3D keypoints triangulated from the manual annotation. The MPJPE is \qty{30.60}{mm} (SE=$\qty{2.03}{mm}$) and the PA-MPJPE is \qty{11.30}{mm} (SE=$\qty{0.80}{mm}$). These are smaller than the estimation accuracies of the state-of-the-art methods and the \METHODNAME{}, as we show next.
We also evaluated the inter-frame consistency of the pose annotations by calculating the angular changes of the joints between consecutive frames in the dataset. The velocity is \ang{3.96} per frame on average, which is comparable to those of the existing datasets (\eg, \ang{2.65} per frame of ARCTIC~\cite{fan2023arctic}).

\subsection{Accuracy of Hand Pose Estimation}
We compare the accuracy of \METHODNAME{} with various baselines fine-tuned on our dataset:
single-image models (HaMeR~\cite{hamer}, WiLoR~\cite{wilor2024} and ArcticNet-SF~\cite{fan2023arctic}), a multi-frame model (ArcticNet-LSTM~\cite{fan2023arctic}), and Body2Hands~\cite{body2hands2021}, which incorporates 3D upper-body keypoints estimated via a monocular estimator~\cite{joo2020eft,rong2021frankmocap}.
For fair comparison, all baselines are fine-tuned for the same iterations as our second stage;
specifically, ArcticNet variants omit the object pose decoder, and
Body2Hands is fine-tuned on our dataset matching the output format to the MANO~\cite{mano2017} parameters.

\Cref{tab:baseline_comparison} shows the result of the evaluation on our dataset.
Our \METHODNAME{} outperforms all of the baseline methods and achieves significantly higher accuracy in joint angle estimation, which indicates that our method can estimate the hand pose from afar more accurately.

\begin{table}[t]
    \centering
    \SetTblrInner{rowsep=2pt,colsep=6pt,stretch=1,abovesep=2pt}
    \begin{tblr}{@{}lrr@{}}
        \toprule
        Method                              & PA-MPJPE (mm)  & Joint Angles (deg) \\
        \midrule
        HaMeR~\cite{hamer}                  & 16.01          & 24.31              \\
        WiLoR~\cite{wilor2024}              & 16.03          & 24.38              \\
        ArcticNet-SF~\cite{fan2023arctic}   & 15.53          & 17.22              \\
        ArcticNet-LSTM~\cite{fan2023arctic} & 14.88          & 16.35              \\
        \midrule
        Body2Hands~\cite{body2hands2021}    & 17.12          & 17.30              \\
        \midrule
        \METHODNAME{} (Ours)                & \textbf{14.47} & \textbf{14.48}     \\
        \bottomrule
    \end{tblr}
    \caption{Comparison of hand pose evaluation metrics between \METHODNAME{} and baselines. \METHODNAME{} outperforms all the baselines in both metrics.}
    \label{tab:baseline_comparison}
\end{table}

\subsection{Qualitative Results}
\begin{figure*}[t]
    \centering
    \begin{tikzpicture}[x=1mm,y=1mm,every node/.style={inner sep=0pt, text depth=0pt}]
        \newcount\colwidth \colwidth=29  
        \newcount\rowheight \rowheight=21 
        \newcount\inlaymargin \inlaymargin=4pt  
        \newcount\examplesmargin \examplesmargin=1.8cm 

        \newcommand\imagewidth{0.16\textwidth} 
        \newcommand\inlaywidth{0.05\textwidth} 

        \newcount\inputcol \inputcol=0
        \newcount\hamercol \hamercol=1
        \newcount\wilorcol \wilorcol=2
        \newcount\arcticsfcol \arcticsfcol=3
        \newcount\proposedcol \proposedcol=4
        \newcount\gtcol \gtcol=5

        \node[font=\bfseries, anchor=south] at (\inputcol*\colwidth,    0) {Input};
        \node[font=\bfseries, anchor=south] at (\proposedcol*\colwidth, 0) {Proposed};
        \node[font=\bfseries, anchor=south] at (\hamercol*\colwidth,    0) {HaMeR};
        \node[font=\bfseries, anchor=south] at (\wilorcol*\colwidth,    0) {WiLoR};
        \node[font=\bfseries, anchor=south] at (\arcticsfcol*\colwidth, 0) {ArcticNet-SF};
        \node[font=\bfseries, anchor=south] at (\gtcol*\colwidth,       0) {GT};

        \newcount\vx \vx=0
        \newcount\vy \vy=0
        \newcount\inlayvy

        \vy = -1
        \node[anchor=north] at (\inputcol*\colwidth,    \vy) {\includegraphics[width=\imagewidth]{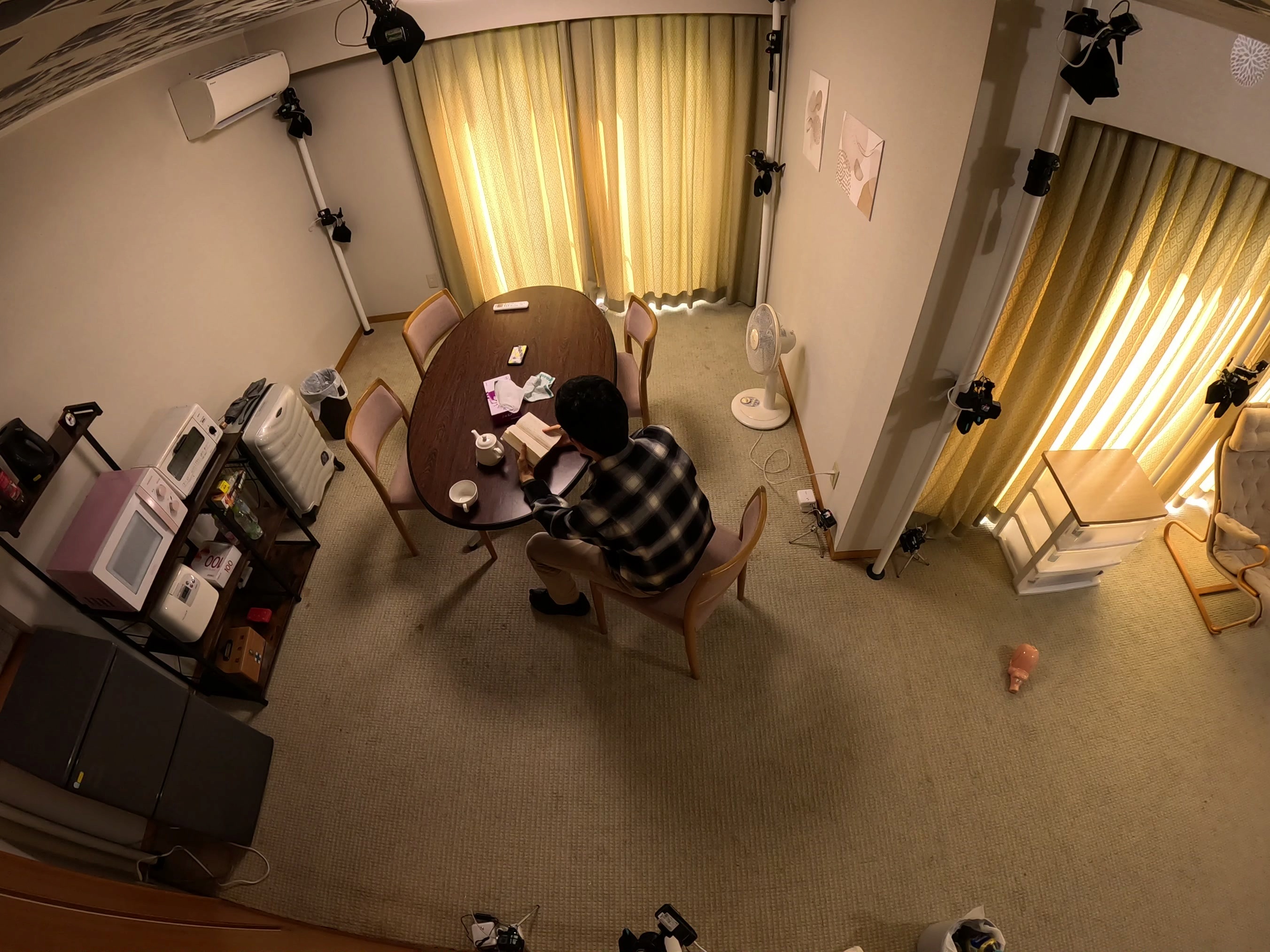}};
        \node[anchor=north] at (\hamercol*\colwidth,    \vy) {\includegraphics[width=\imagewidth]{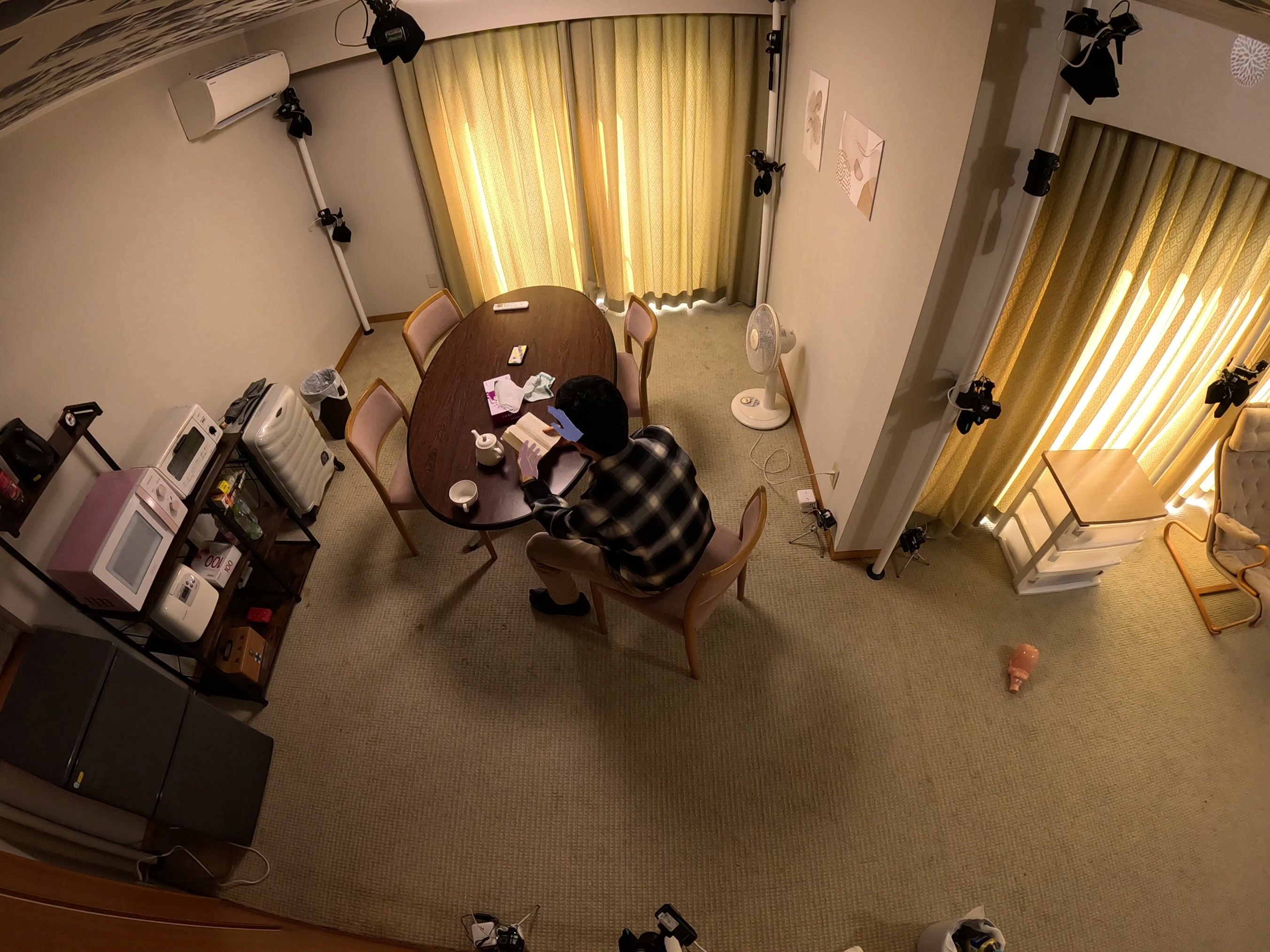}};
        \node[anchor=north] at (\wilorcol*\colwidth,    \vy) {\includegraphics[width=\imagewidth]{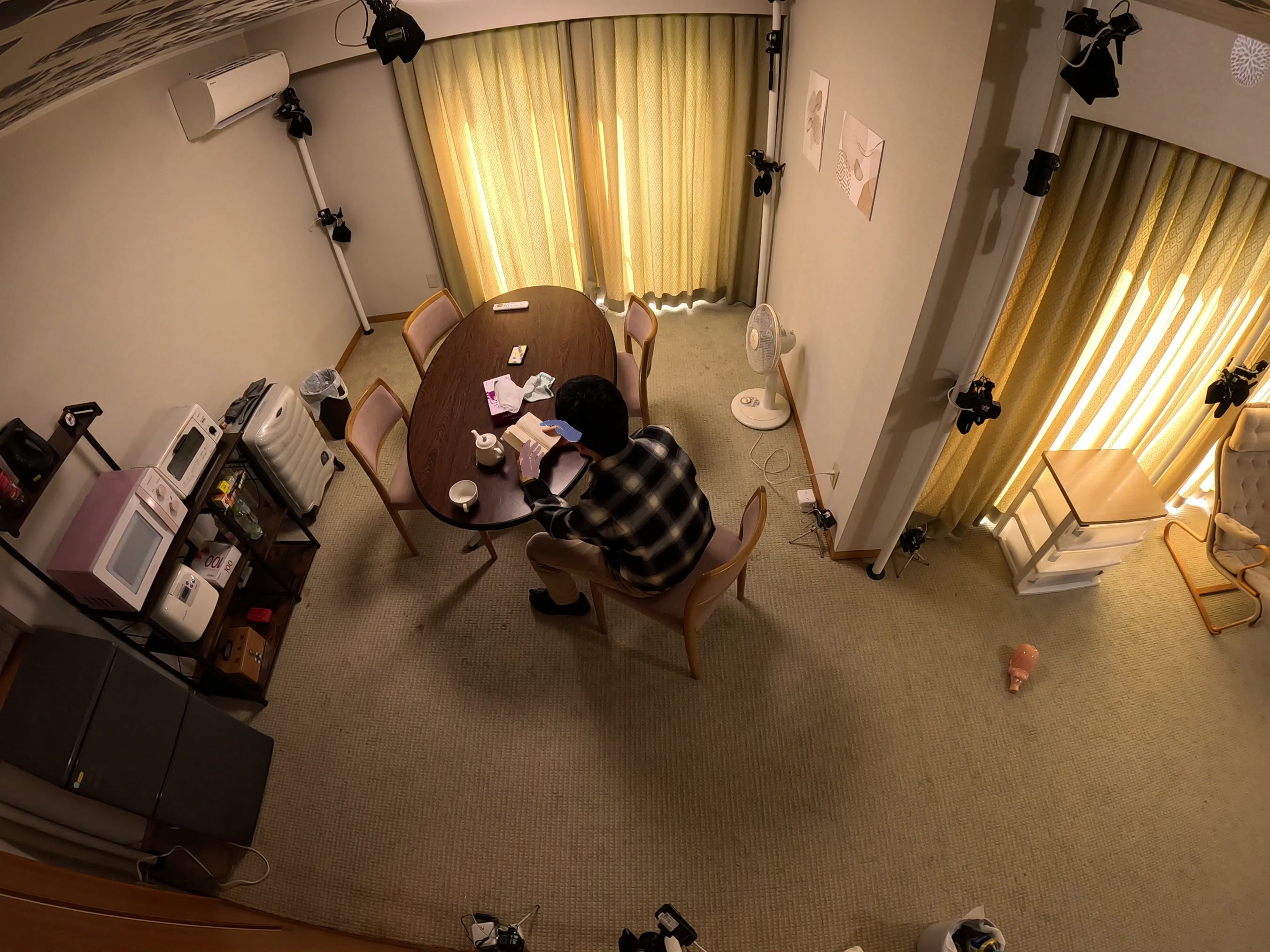}};
        \node[anchor=north] at (\arcticsfcol*\colwidth, \vy) {\includegraphics[width=\imagewidth]{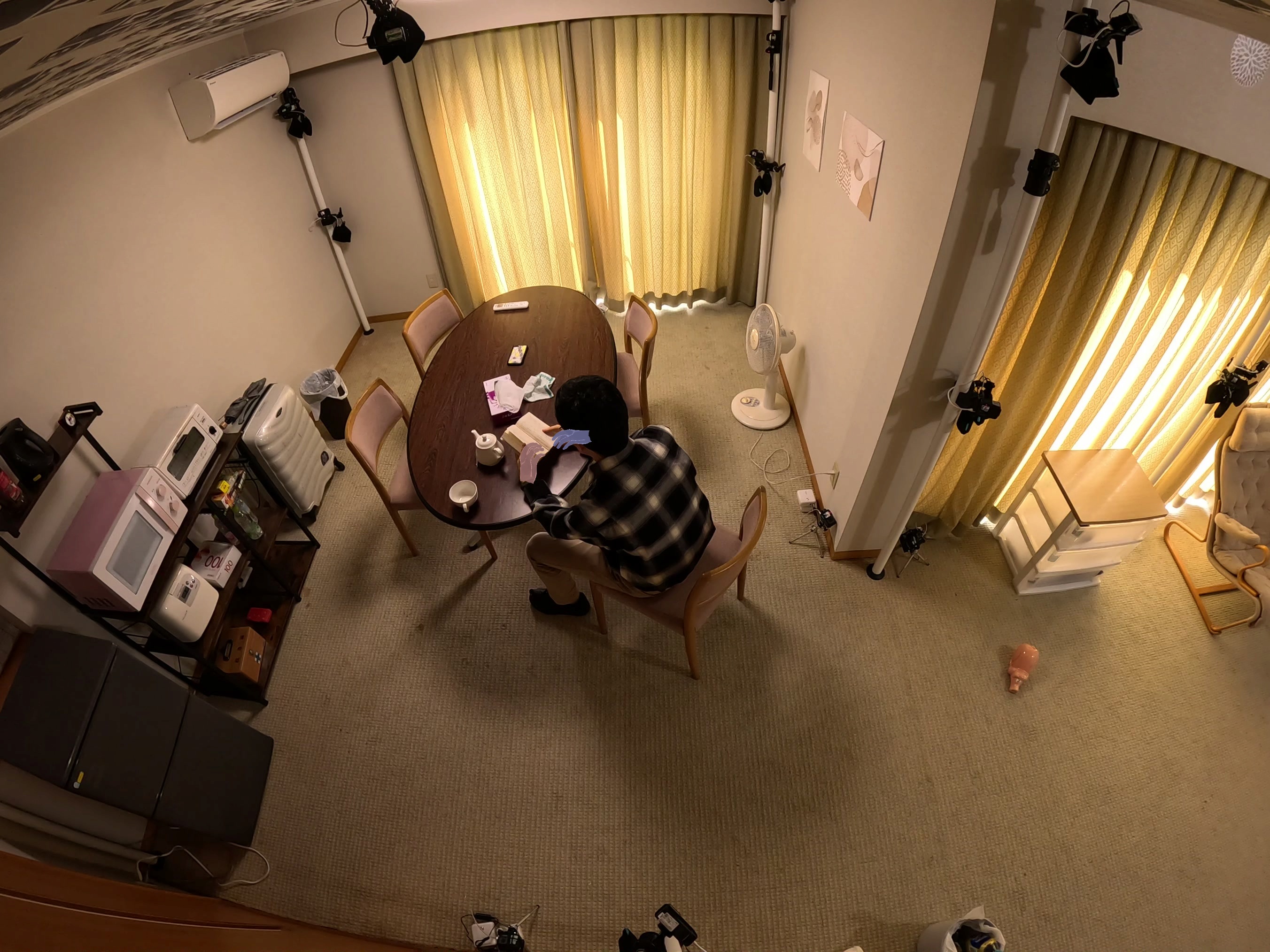}};
        \node[anchor=north] at (\gtcol*\colwidth,       \vy) {\includegraphics[width=\imagewidth]{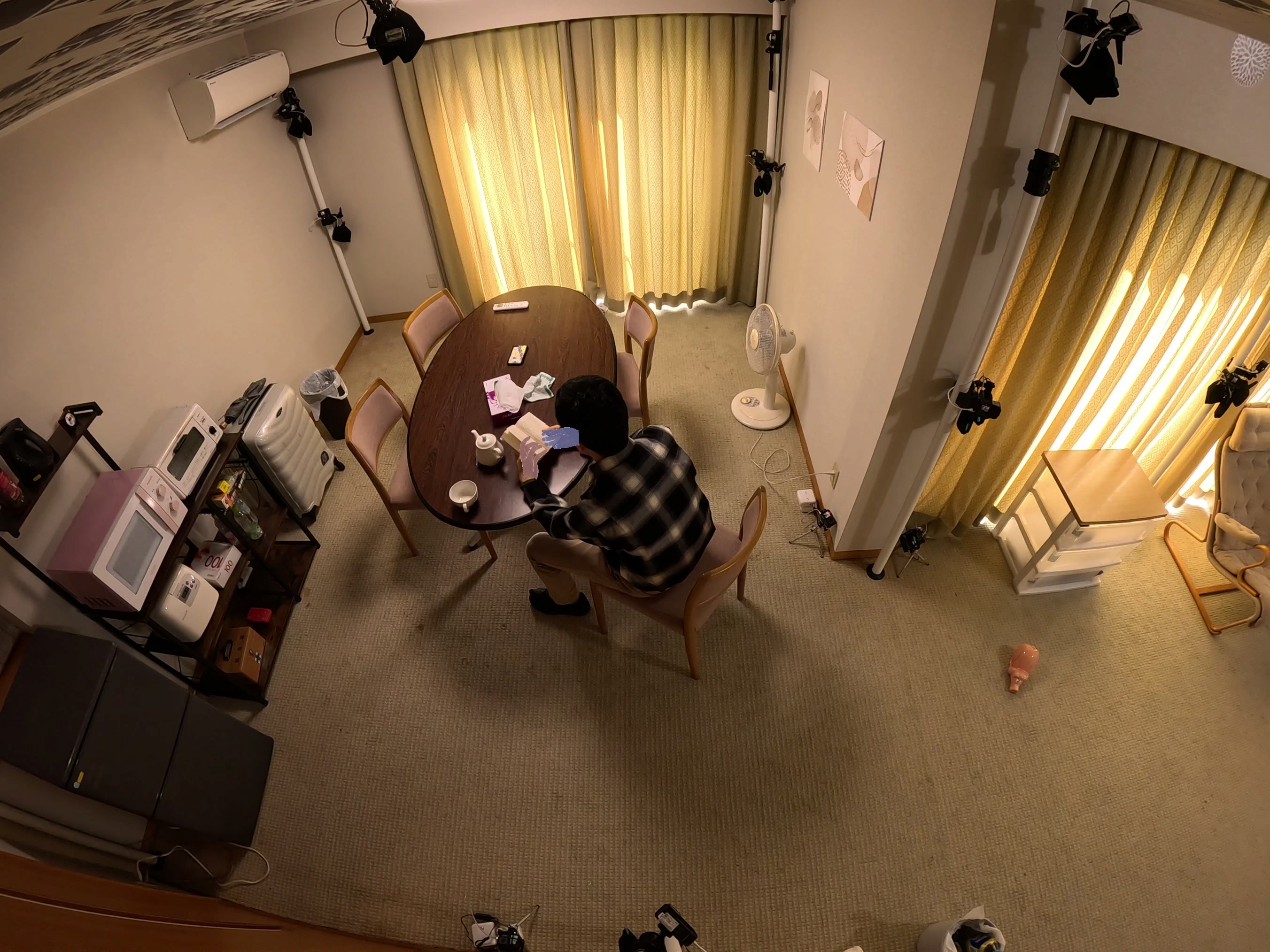}};
        \node[anchor=north] at (\proposedcol*\colwidth, \vy) {\includegraphics[width=\imagewidth]{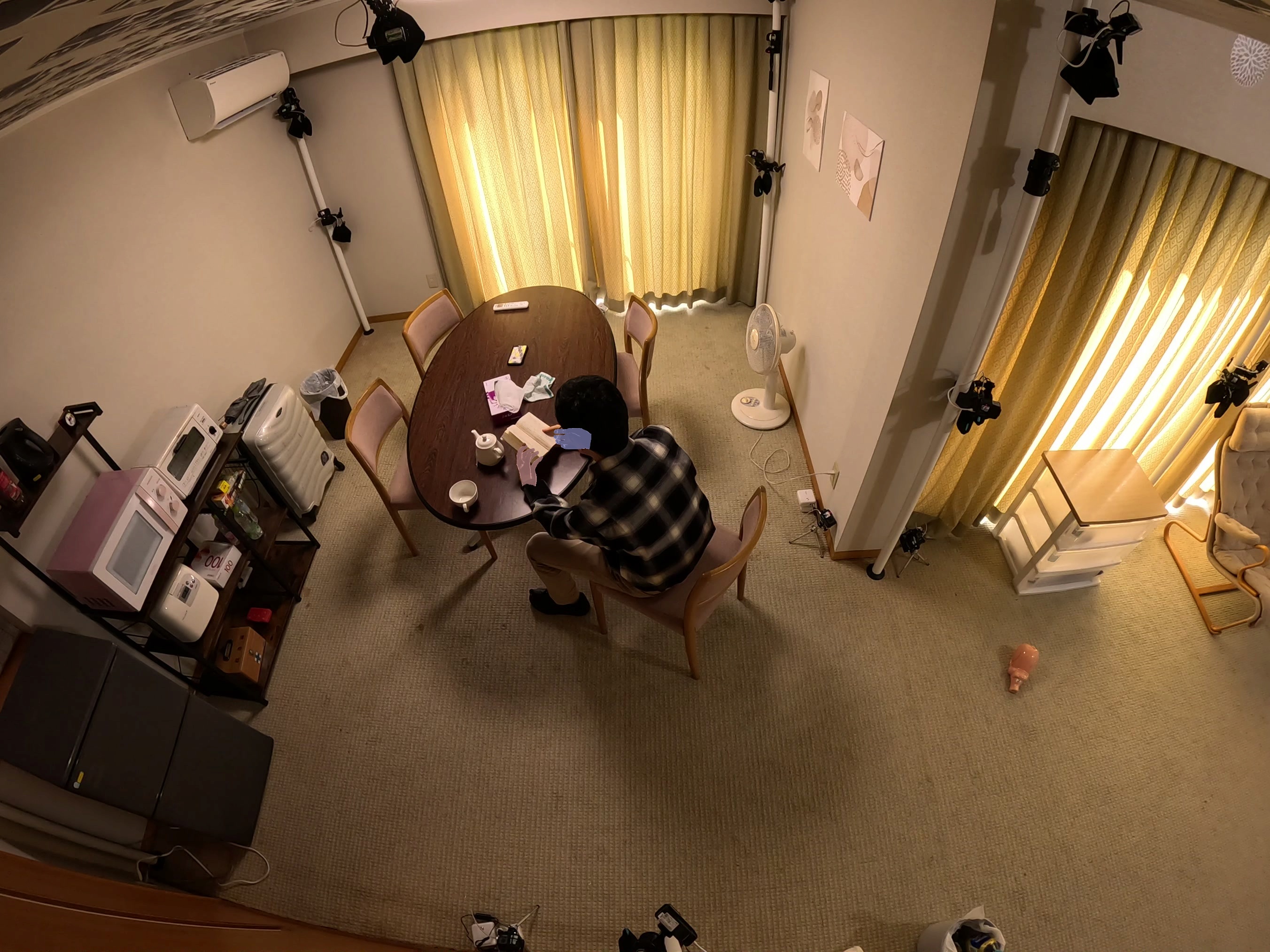}};

        \vx = \colwidth
        \advance \vx by 8
        \inlayvy = \vy
        \advance \inlayvy by -9
        \node (inlay_hamer_right) [inner sep=0pt, draw=none, fill=white, fill opacity=0.3, minimum width=\inlaywidth+\inlaymargin, minimum height=\inlaywidth+\inlaymargin] at (\vx, \inlayvy )  {};
        \node[anchor=center] at (inlay_hamer_right.center) {\includegraphics[width=\inlaywidth]{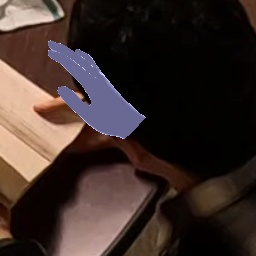}};
        \advance\vx by \colwidth
        \node (inlay_wilor_right) [inner sep=0pt, draw=none, fill=white, fill opacity=0.3, minimum width=\inlaywidth+\inlaymargin, minimum height=\inlaywidth+\inlaymargin] at (\vx, \inlayvy )  {};
        \node[anchor=center] at (inlay_wilor_right.center) {\includegraphics[width=\inlaywidth]{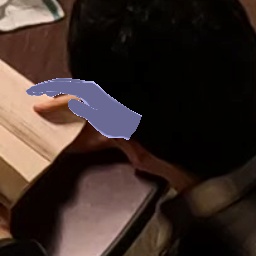}};
        \advance\vx by \colwidth
        \node (inlay_arcticsf_right) [inner sep=0pt, draw=none, fill=white, fill opacity=0.3, minimum width=\inlaywidth+\inlaymargin, minimum height=\inlaywidth+\inlaymargin] at (\vx, \inlayvy )  {};
        \node[anchor=center] at (inlay_arcticsf_right.center) {\includegraphics[width=\inlaywidth]{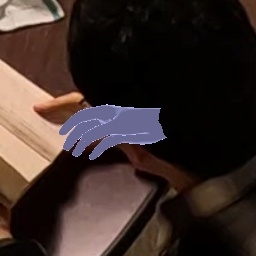}};
        \advance\vx by \colwidth
        \node (inlay_proposed_right) [inner sep=0pt, draw=none, fill=white, fill opacity=0.3, minimum width=\inlaywidth+\inlaymargin, minimum height=\inlaywidth+\inlaymargin] at (\vx, \inlayvy )  {};
        \node[anchor=center] at (inlay_proposed_right.center) {\includegraphics[width=\inlaywidth]{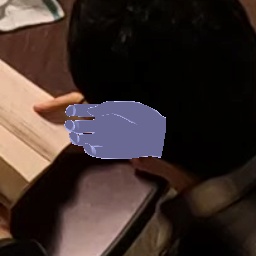}};
        \advance\vx by \colwidth
        \node (inlay_gt_right) [inner sep=0pt, draw=none, fill=white, fill opacity=0.3, minimum width=\inlaywidth+\inlaymargin, minimum height=\inlaywidth+\inlaymargin] at (\vx, \inlayvy )  {};
        \node[anchor=center] at (inlay_gt_right.center) {\includegraphics[width=\inlaywidth]{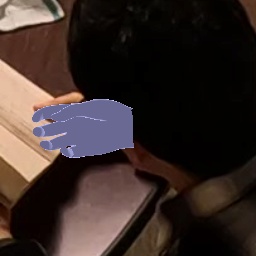}};

        \vx = \colwidth
        \advance \vx by -8
        \inlayvy = \vy
        \advance \inlayvy by -15
        \node (inlay_hamer_left) [inner sep=0pt, draw=none, fill=white, fill opacity=0.3, minimum width=\inlaywidth+\inlaymargin, minimum height=\inlaywidth+\inlaymargin] at (\vx, \inlayvy )  {};
        \node[anchor=center] at (inlay_hamer_left.center) {\includegraphics[width=\inlaywidth]{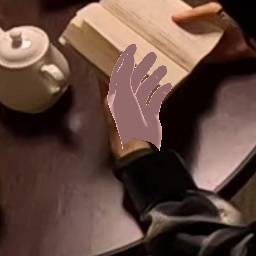}};
        \advance\vx by \colwidth
        \node (inlay_wilor_left) [inner sep=0pt, draw=none, fill=white, fill opacity=0.3, minimum width=\inlaywidth+\inlaymargin, minimum height=\inlaywidth+\inlaymargin] at (\vx, \inlayvy )  {};
        \node[anchor=center] at (inlay_wilor_left.center) {\includegraphics[width=\inlaywidth]{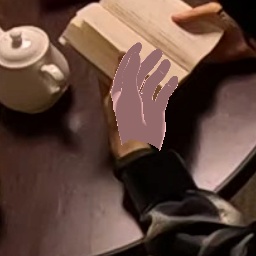}};
        \advance\vx by \colwidth
        \node (inlay_arcticsf_left) [inner sep=0pt, draw=none, fill=white, fill opacity=0.3, minimum width=\inlaywidth+\inlaymargin, minimum height=\inlaywidth+\inlaymargin] at (\vx, \inlayvy )  {};
        \node[anchor=center] at (inlay_arcticsf_left.center) {\includegraphics[width=\inlaywidth]{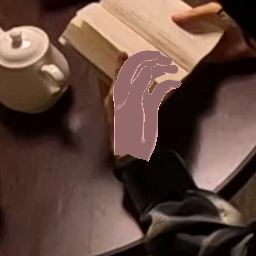}};
        \advance\vx by \colwidth
        \node (inlay_proposed_left) [inner sep=0pt, draw=none, fill=white, fill opacity=0.3, minimum width=\inlaywidth+\inlaymargin, minimum height=\inlaywidth+\inlaymargin] at (\vx, \inlayvy )  {};
        \node[anchor=center] at (inlay_proposed_left.center) {\includegraphics[width=\inlaywidth]{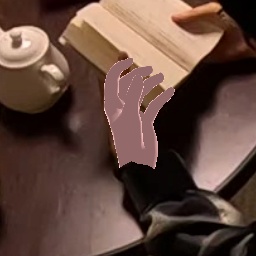}};
        \advance\vx by \colwidth
        \node (inlay_gt_left) [inner sep=0pt, draw=none, fill=white, fill opacity=0.3, minimum width=\inlaywidth+\inlaymargin, minimum height=\inlaywidth+\inlaymargin] at (\vx, \inlayvy )  {};
        \node[anchor=center] at (inlay_gt_left.center) {\includegraphics[width=\inlaywidth]{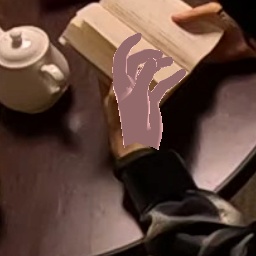}};

        \advance\vy by -\rowheight
        \node[anchor=north] at (\inputcol*\colwidth,    \vy) {\includegraphics[width=\imagewidth]{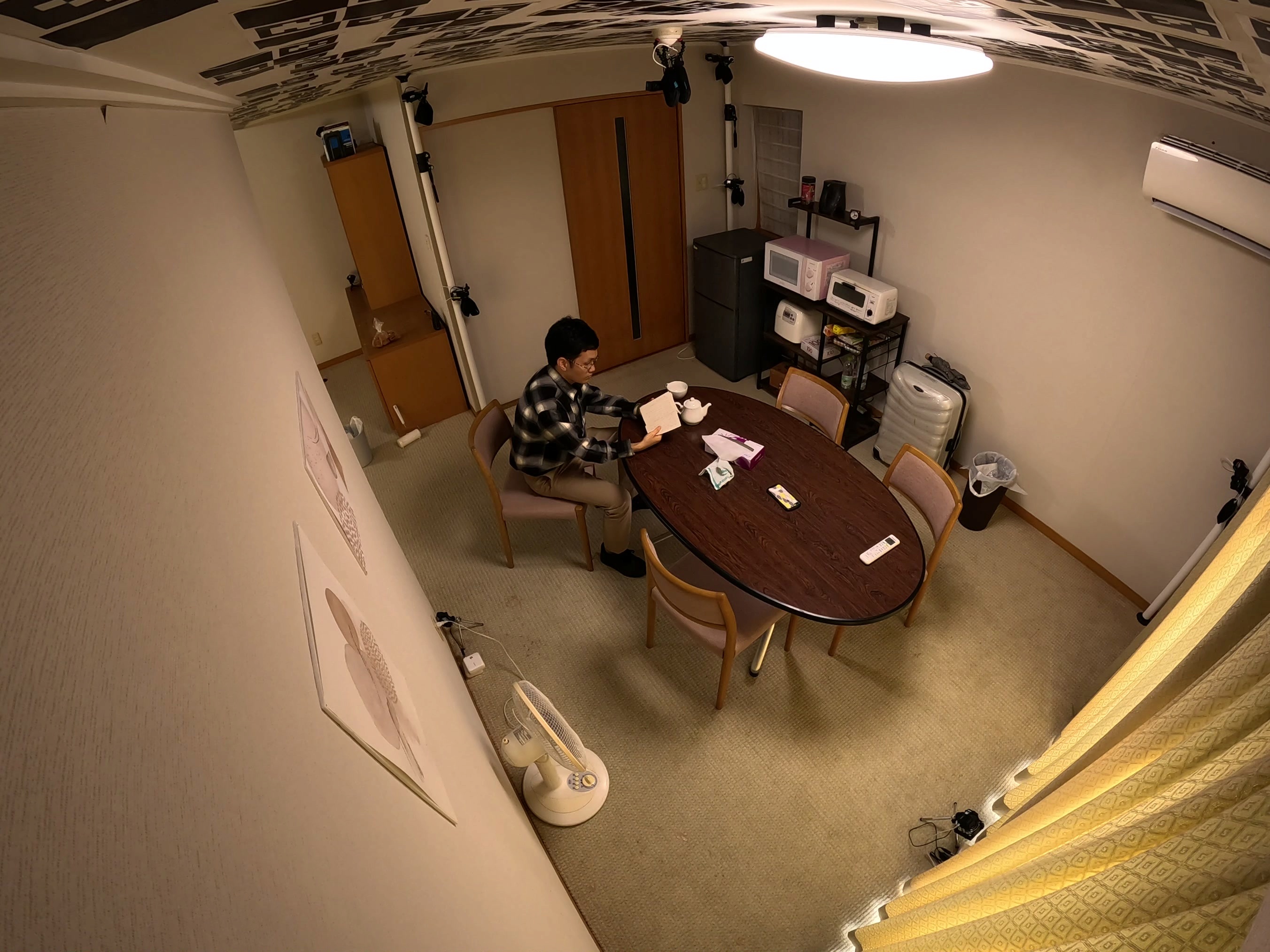}};
        \node[anchor=north] at (\proposedcol*\colwidth, \vy) {\includegraphics[width=\imagewidth]{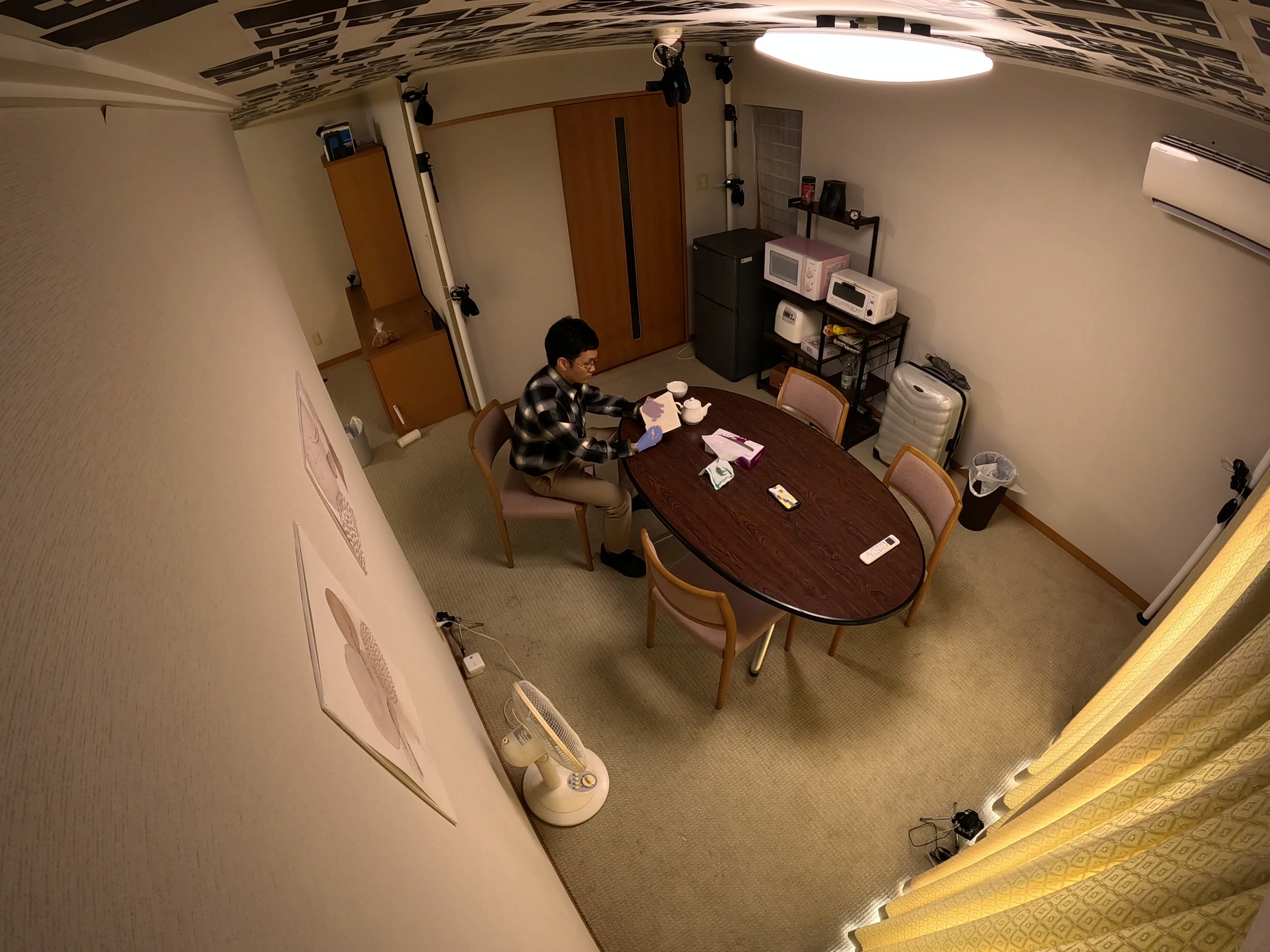}};
        \node[anchor=north] at (\hamercol*\colwidth,    \vy) {\includegraphics[width=\imagewidth]{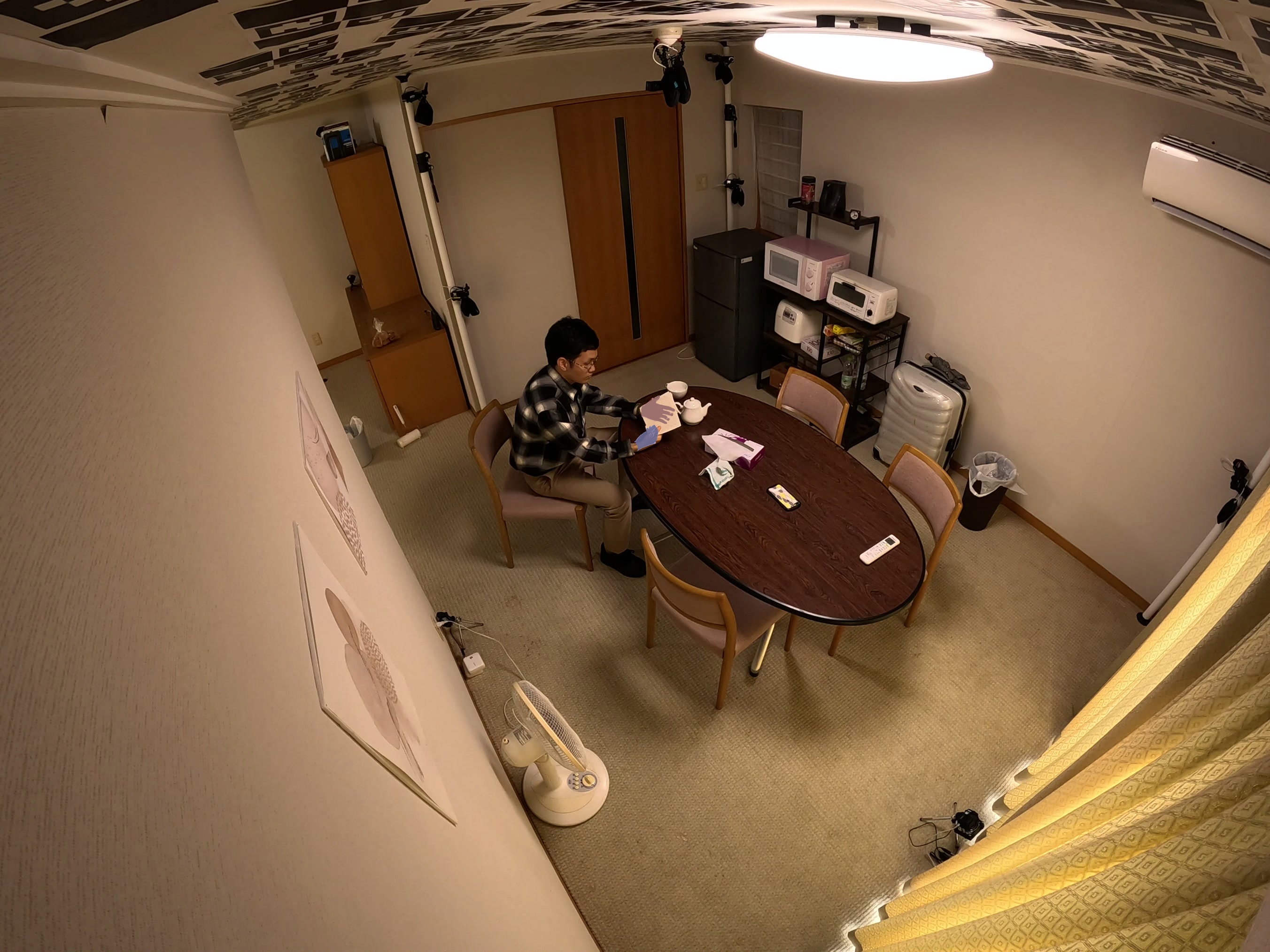}};
        \node[anchor=north] at (\wilorcol*\colwidth,    \vy) {\includegraphics[width=\imagewidth]{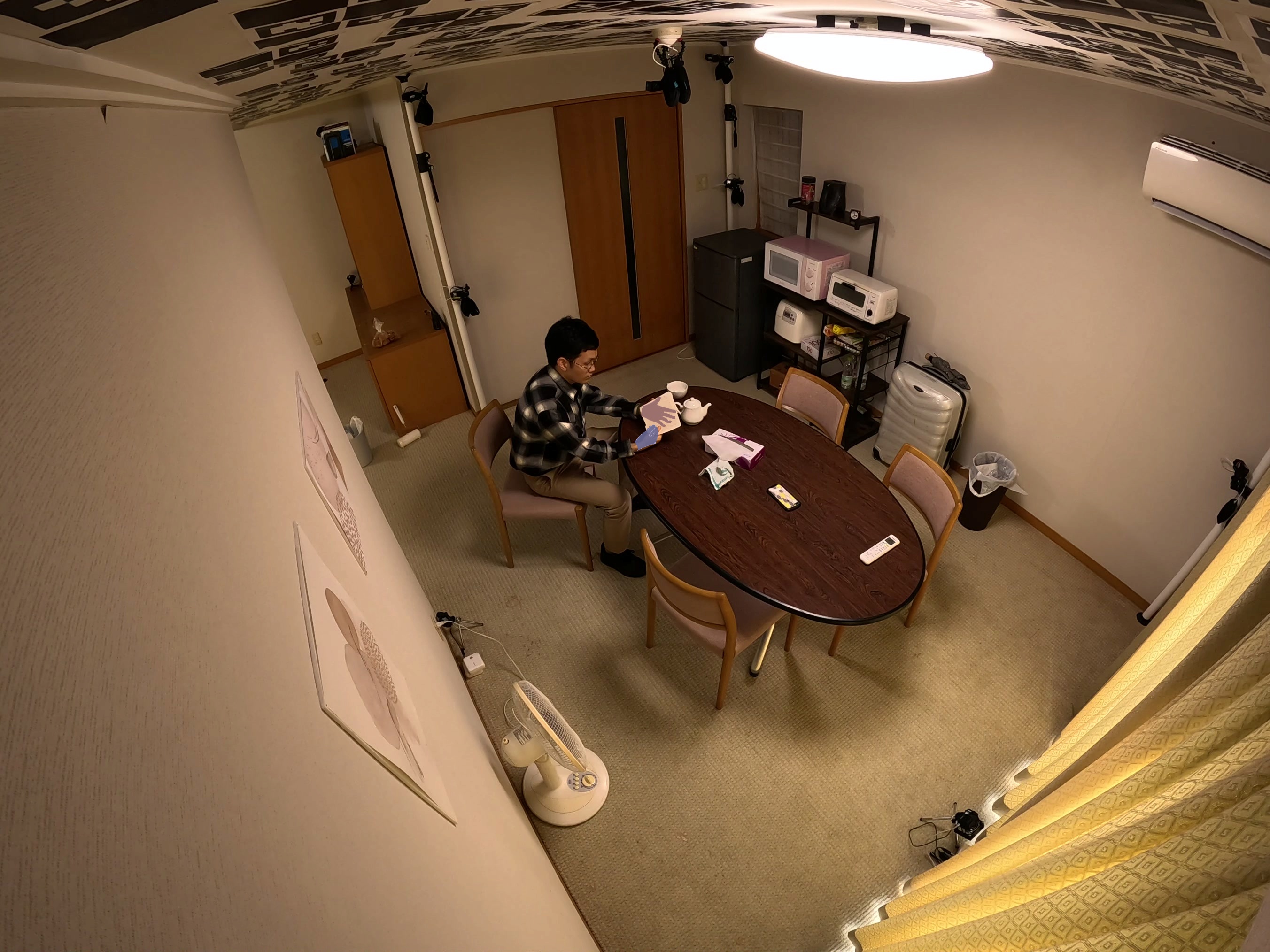}};
        \node[anchor=north] at (\arcticsfcol*\colwidth, \vy) {\includegraphics[width=\imagewidth]{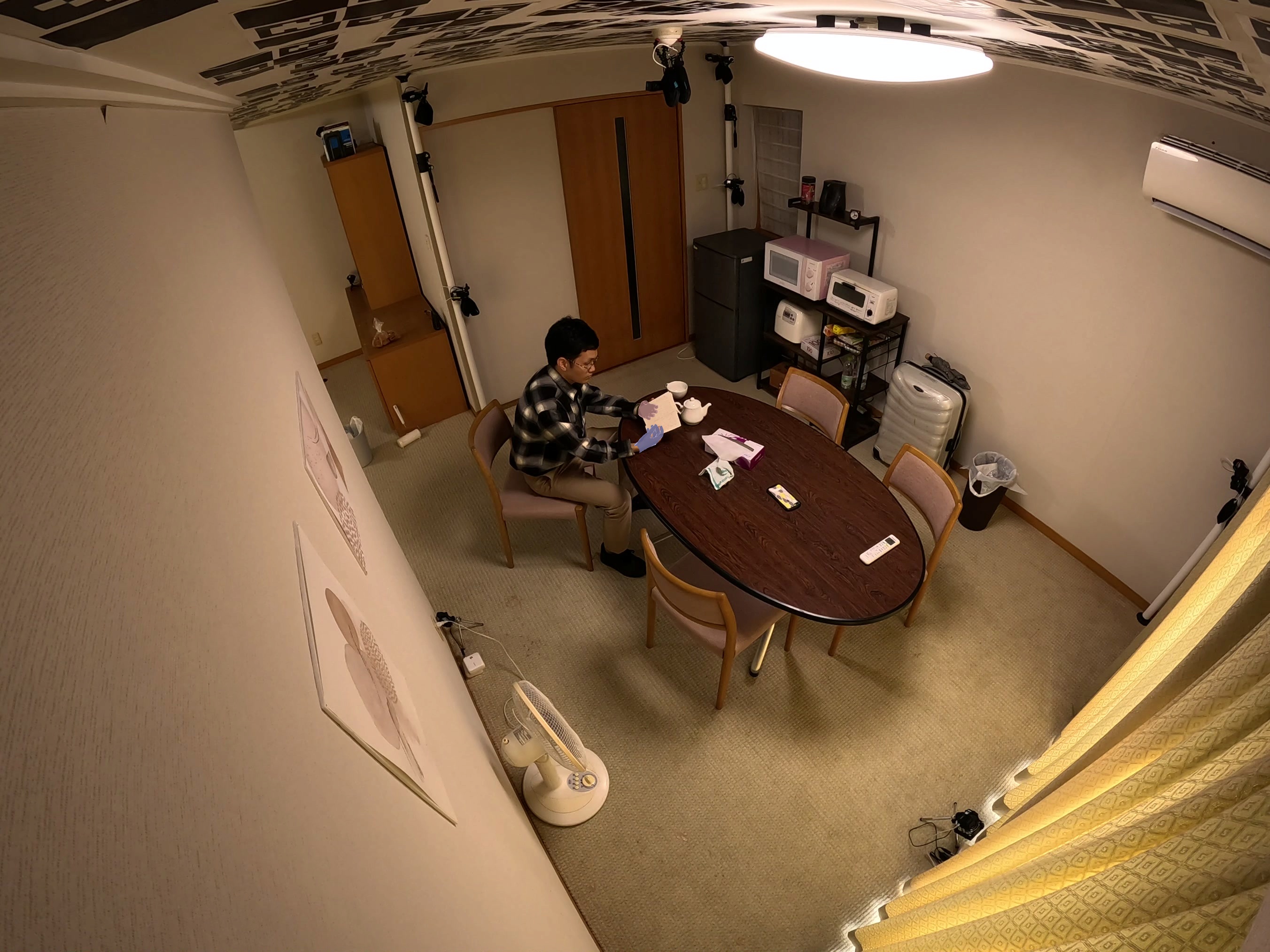}};
        \node[anchor=north] at (\gtcol*\colwidth,       \vy) {\includegraphics[width=\imagewidth]{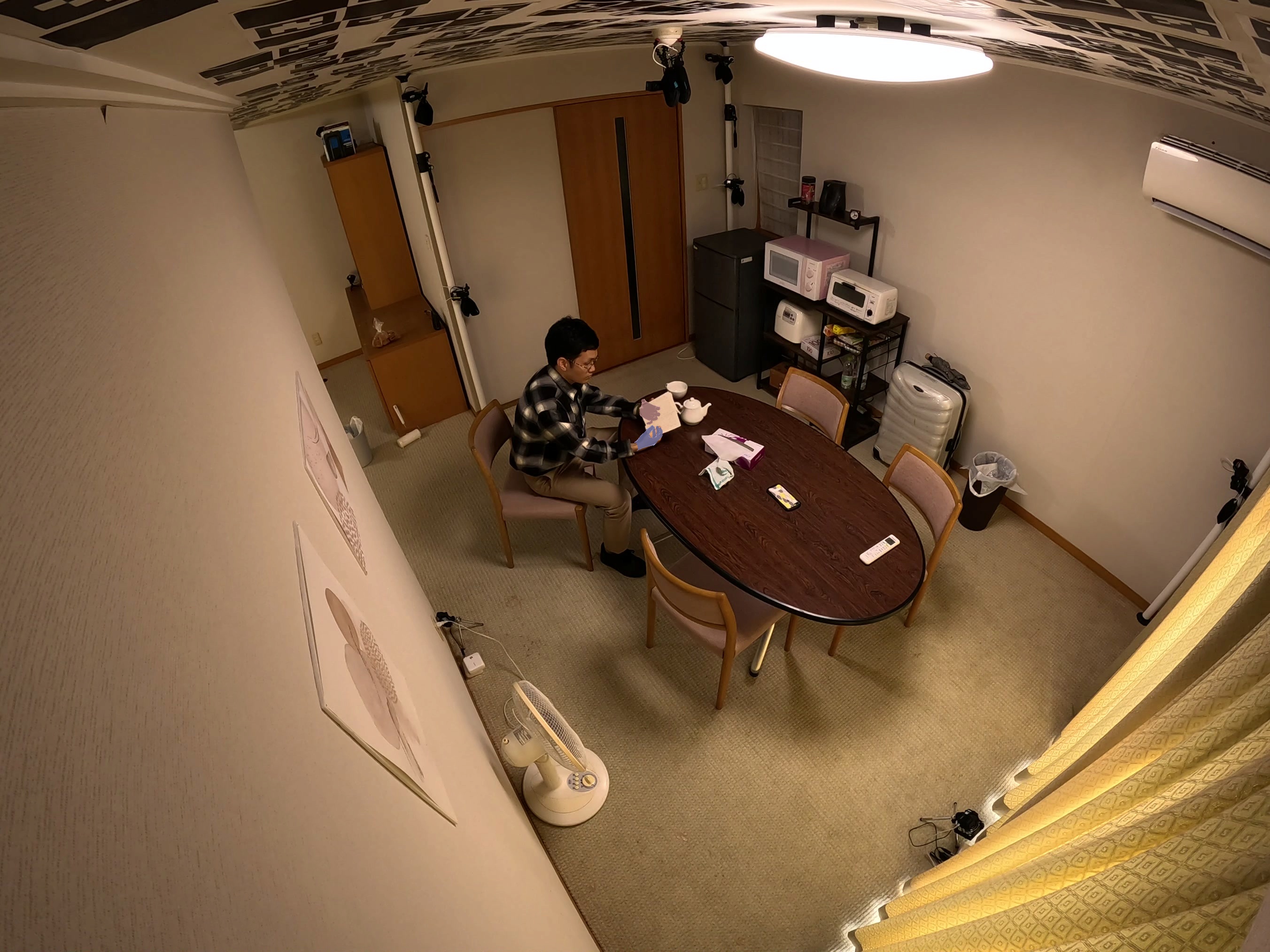}};

        \vx = \colwidth
        \advance \vx by 8
        \inlayvy = \vy
        \advance \inlayvy by -9
        \node (inlay_hamer_right) [inner sep=0pt, draw=none, fill=white, fill opacity=0.3, minimum width=\inlaywidth+\inlaymargin, minimum height=\inlaywidth+\inlaymargin] at (\vx, \inlayvy )  {};
        \node[anchor=center] at (inlay_hamer_right.center) {\includegraphics[width=\inlaywidth]{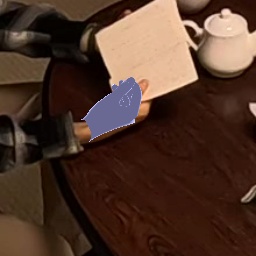}};
        \advance\vx by \colwidth
        \node (inlay_wilor_right) [inner sep=0pt, draw=none, fill=white, fill opacity=0.3, minimum width=\inlaywidth+\inlaymargin, minimum height=\inlaywidth+\inlaymargin] at (\vx, \inlayvy )  {};
        \node[anchor=center] at (inlay_wilor_right.center) {\includegraphics[width=\inlaywidth]{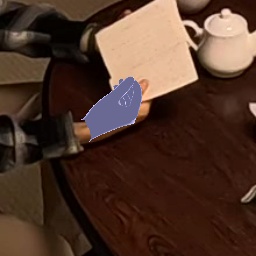}};
        \advance\vx by \colwidth
        \node (inlay_arcticsf_right) [inner sep=0pt, draw=none, fill=white, fill opacity=0.3, minimum width=\inlaywidth+\inlaymargin, minimum height=\inlaywidth+\inlaymargin] at (\vx, \inlayvy )  {};
        \node[anchor=center] at (inlay_arcticsf_right.center) {\includegraphics[width=\inlaywidth]{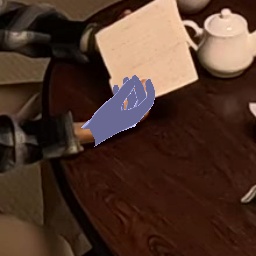}};
        \advance\vx by \colwidth
        \node (inlay_proposed_right) [inner sep=0pt, draw=none, fill=white, fill opacity=0.3, minimum width=\inlaywidth+\inlaymargin, minimum height=\inlaywidth+\inlaymargin] at (\vx, \inlayvy )  {};
        \node[anchor=center] at (inlay_proposed_right.center) {\includegraphics[width=\inlaywidth]{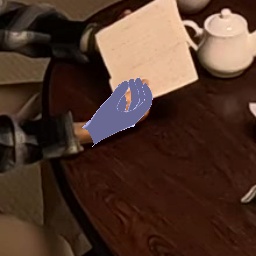}};
        \advance\vx by \colwidth
        \node (inlay_gt_right) [inner sep=0pt, draw=none, fill=white, fill opacity=0.3, minimum width=\inlaywidth+\inlaymargin, minimum height=\inlaywidth+\inlaymargin] at (\vx, \inlayvy )  {};
        \node[anchor=center] at (inlay_gt_right.center) {\includegraphics[width=\inlaywidth]{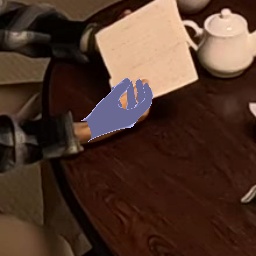}};

        \vx = \colwidth
        \advance \vx by -8
        \inlayvy = \vy
        \advance \inlayvy by -15
        \node (inlay_hamer_left) [inner sep=0pt, draw=none, fill=white, fill opacity=0.3, minimum width=\inlaywidth+\inlaymargin, minimum height=\inlaywidth+\inlaymargin] at (\vx, \inlayvy )  {};
        \node[anchor=center] at (inlay_hamer_left.center) {\includegraphics[width=\inlaywidth]{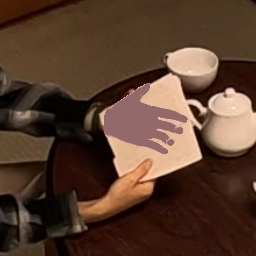}};
        \advance\vx by \colwidth
        \node (inlay_wilor_left) [inner sep=0pt, draw=none, fill=white, fill opacity=0.3, minimum width=\inlaywidth+\inlaymargin, minimum height=\inlaywidth+\inlaymargin] at (\vx, \inlayvy )  {};
        \node[anchor=center] at (inlay_wilor_left.center) {\includegraphics[width=\inlaywidth]{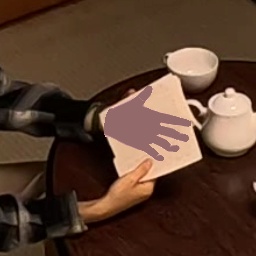}};
        \advance\vx by \colwidth
        \node (inlay_arcticsf_left) [inner sep=0pt, draw=none, fill=white, fill opacity=0.3, minimum width=\inlaywidth+\inlaymargin, minimum height=\inlaywidth+\inlaymargin] at (\vx, \inlayvy )  {};
        \node[anchor=center] at (inlay_arcticsf_left.center) {\includegraphics[width=\inlaywidth]{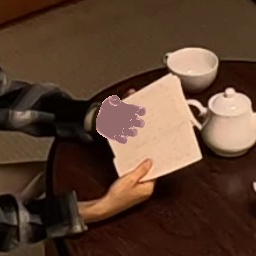}};
        \advance\vx by \colwidth
        \node (inlay_proposed_left) [inner sep=0pt, draw=none, fill=white, fill opacity=0.3, minimum width=\inlaywidth+\inlaymargin, minimum height=\inlaywidth+\inlaymargin] at (\vx, \inlayvy )  {};
        \node[anchor=center] at (inlay_proposed_left.center) {\includegraphics[width=\inlaywidth]{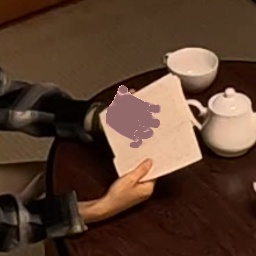}};
        \advance\vx by \colwidth
        \node (inlay_gt_left) [inner sep=0pt, draw=none, fill=white, fill opacity=0.3, minimum width=\inlaywidth+\inlaymargin, minimum height=\inlaywidth+\inlaymargin] at (\vx, \inlayvy )  {};
        \node[anchor=center] at (inlay_gt_left.center) {\includegraphics[width=\inlaywidth]{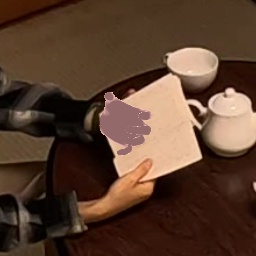}};

        \advance\vy by -\rowheight
        \advance\vy by -\examplesmargin
        \node[anchor=north] at (\inputcol*\colwidth,    \vy) {\includegraphics[width=\imagewidth]{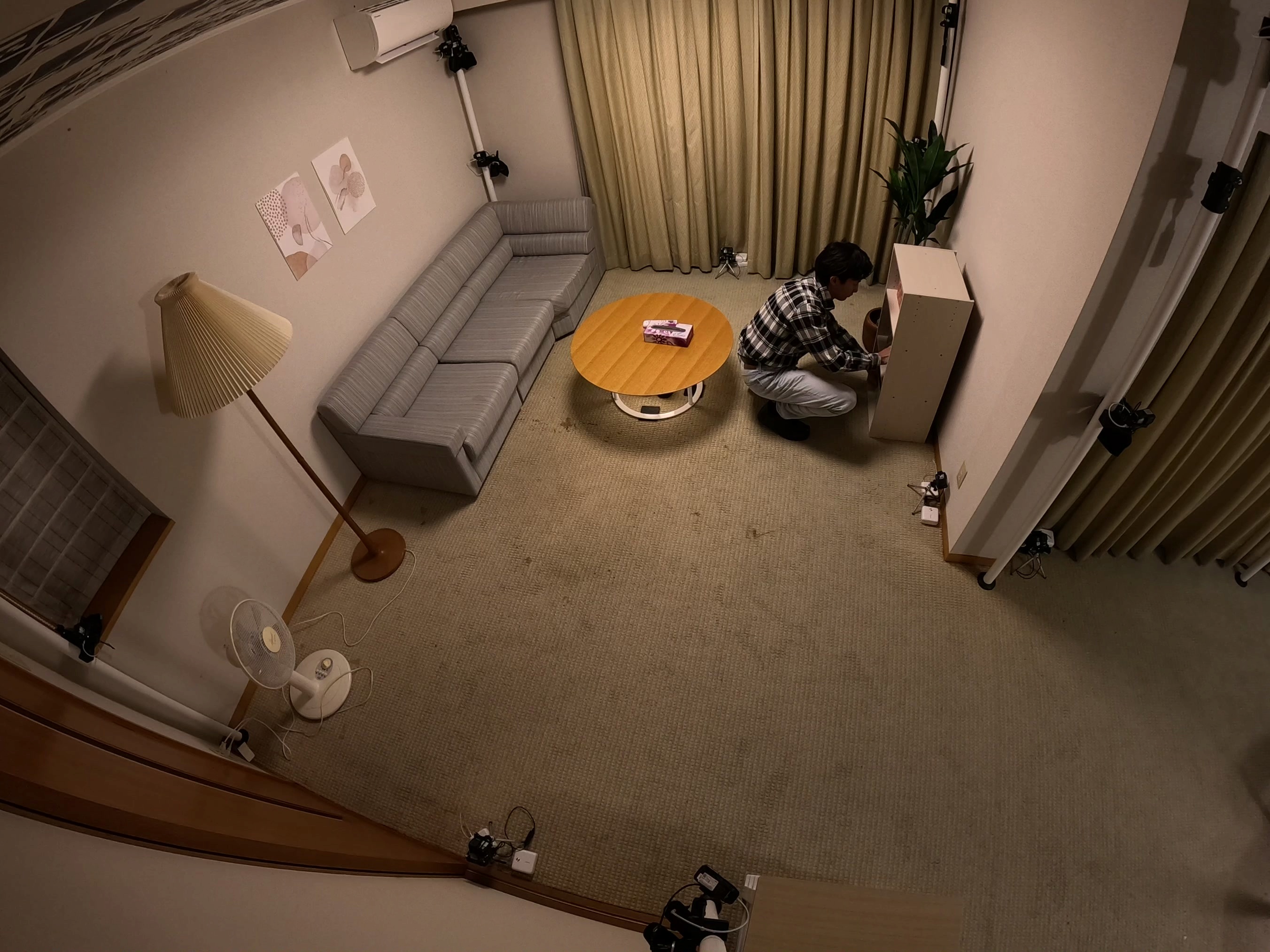}};
        \node[anchor=north] at (\proposedcol*\colwidth, \vy) {\includegraphics[width=\imagewidth]{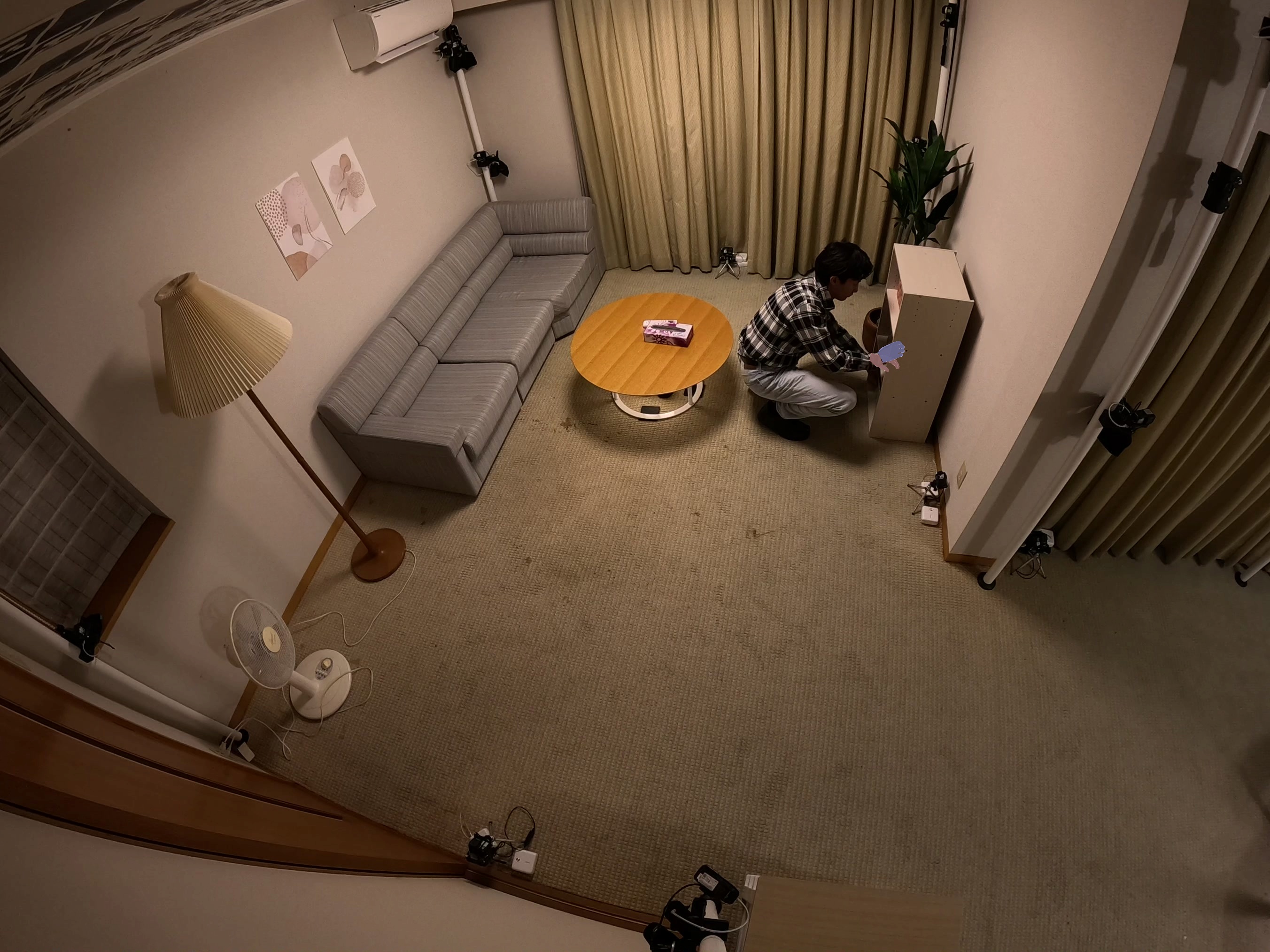}};
        \node[anchor=north] at (\hamercol*\colwidth,    \vy) {\includegraphics[width=\imagewidth]{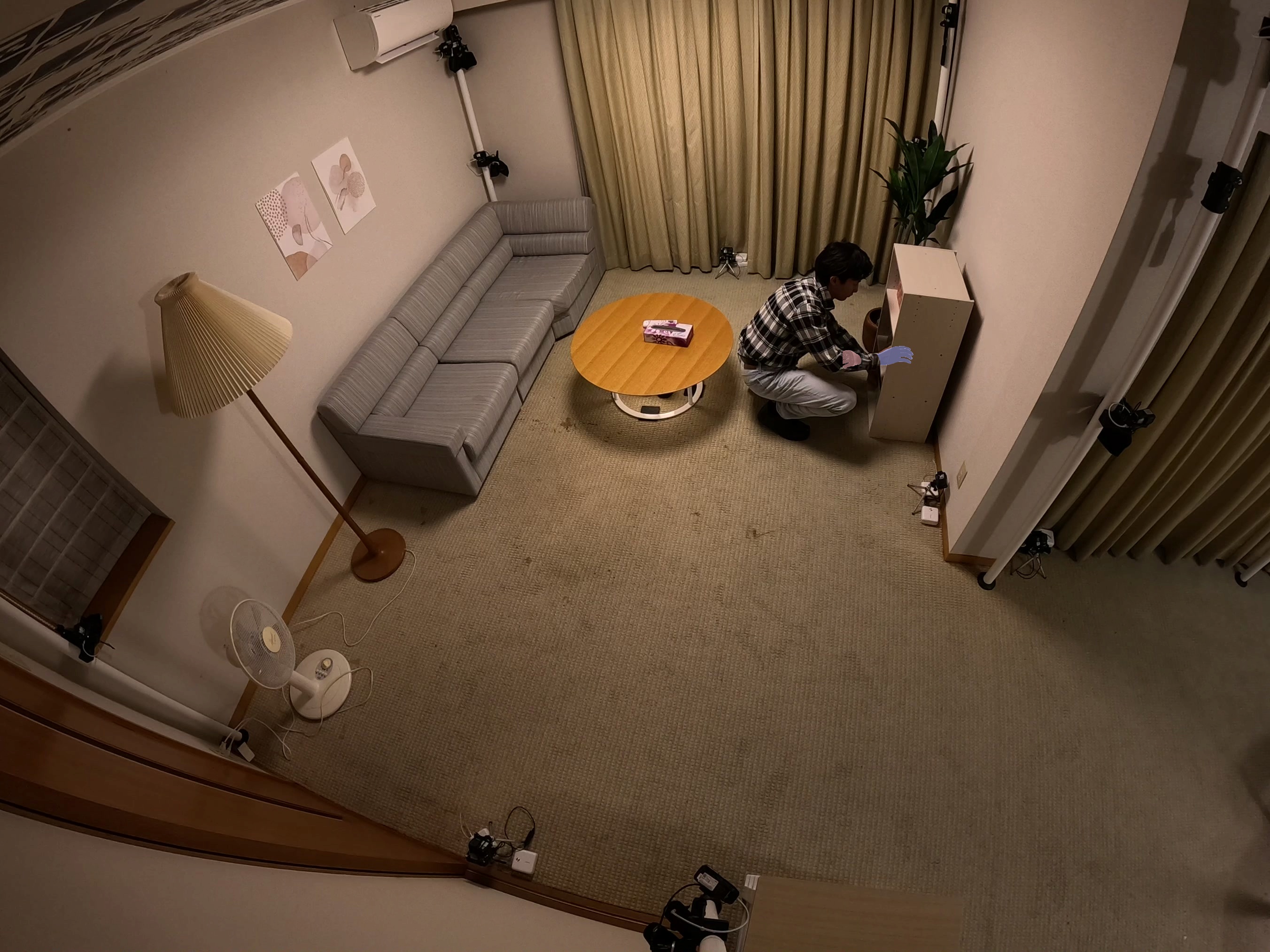}};
        \node[anchor=north] at (\wilorcol*\colwidth,    \vy) {\includegraphics[width=\imagewidth]{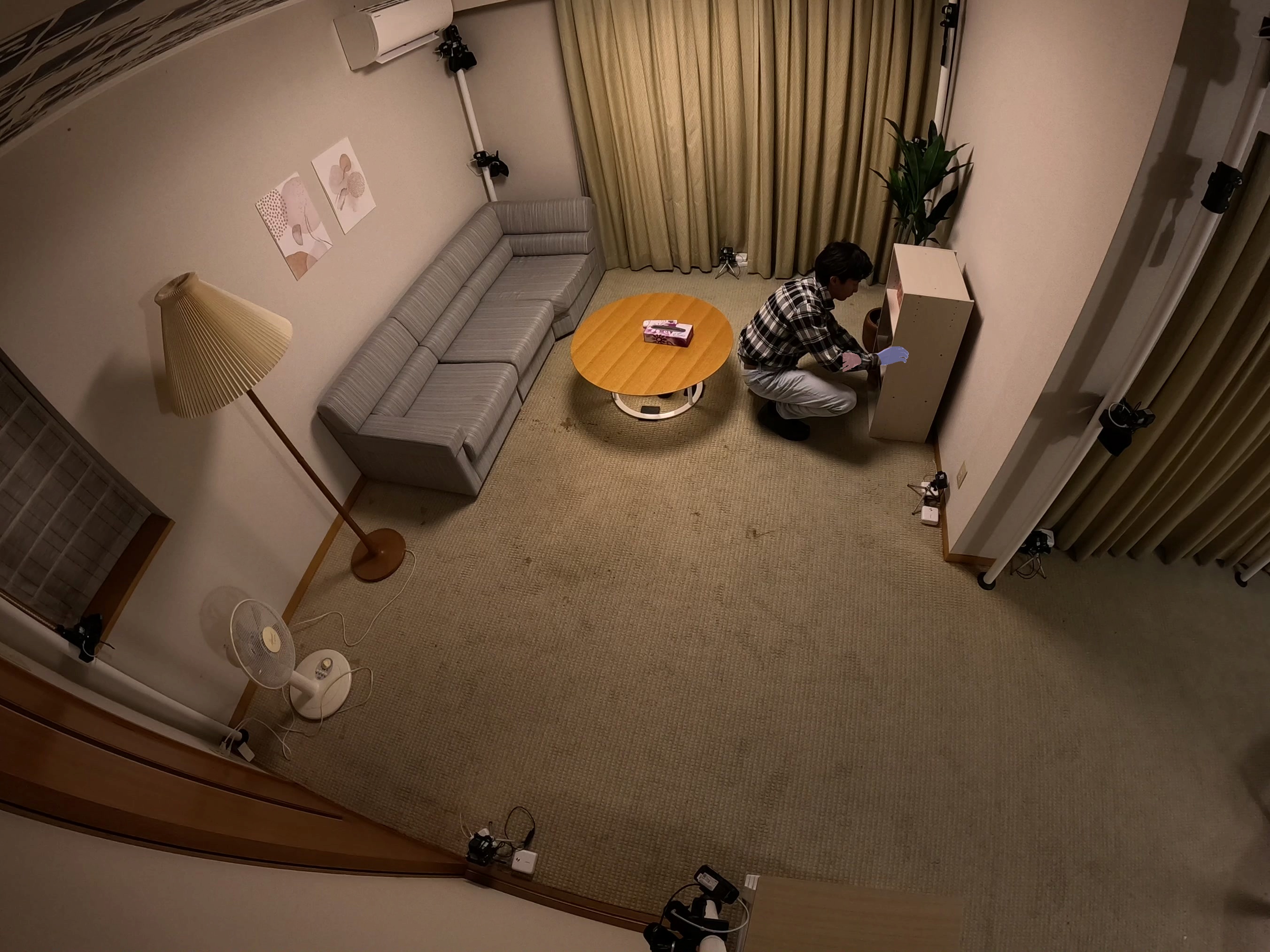}};
        \node[anchor=north] at (\arcticsfcol*\colwidth, \vy) {\includegraphics[width=\imagewidth]{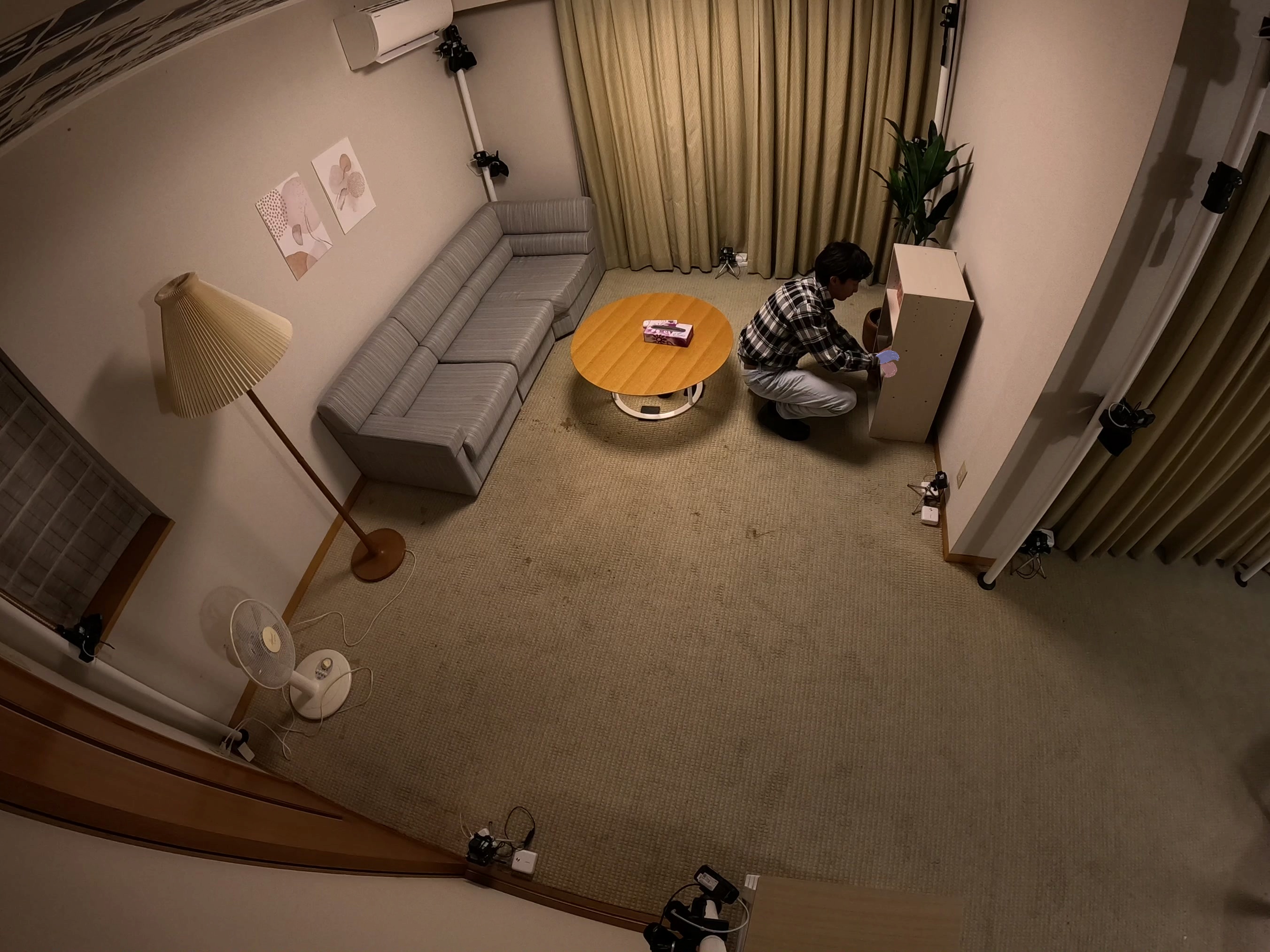}};
        \node[anchor=north] at (\gtcol*\colwidth,       \vy) {\includegraphics[width=\imagewidth]{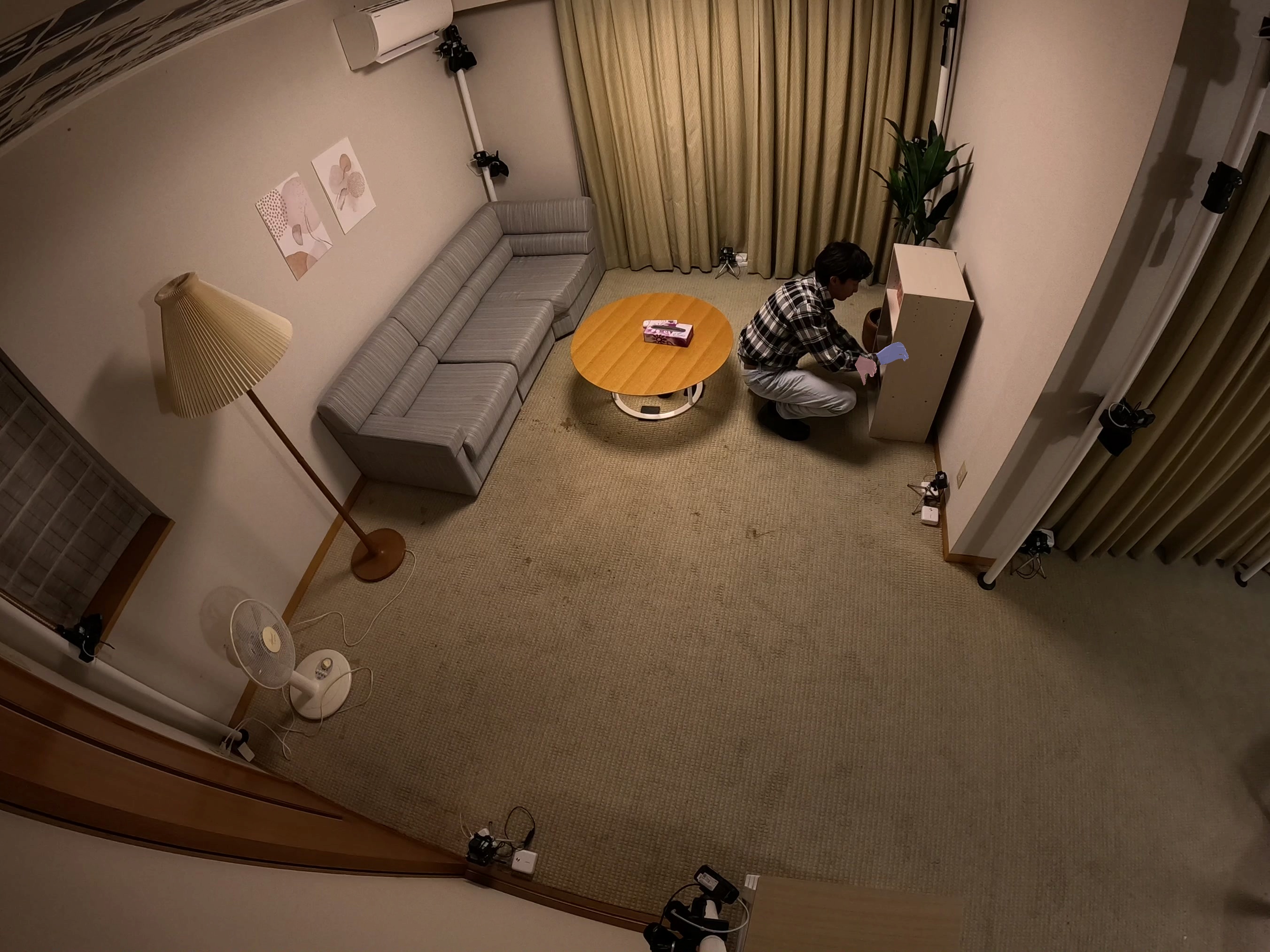}};

        \vx = \colwidth
        \advance \vx by 3
        \inlayvy = \vy
        \advance \inlayvy by -16
        \node (inlay_hamer_right) [inner sep=0pt, draw=none, fill=white, fill opacity=0.3, minimum width=\inlaywidth+\inlaymargin, minimum height=\inlaywidth+\inlaymargin] at (\vx, \inlayvy )  {};
        \node[anchor=center] at (inlay_hamer_right.center) {\includegraphics[width=\inlaywidth]{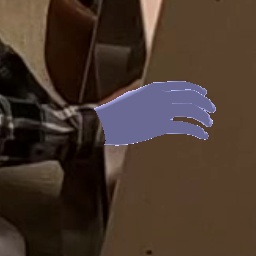}};
        \advance\vx by \colwidth
        \node (inlay_wilor_right) [inner sep=0pt, draw=none, fill=white, fill opacity=0.3, minimum width=\inlaywidth+\inlaymargin, minimum height=\inlaywidth+\inlaymargin] at (\vx, \inlayvy )  {};
        \node[anchor=center] at (inlay_wilor_right.center) {\includegraphics[width=\inlaywidth]{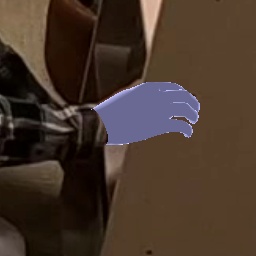}};
        \advance\vx by \colwidth
        \node (inlay_arcticsf_right) [inner sep=0pt, draw=none, fill=white, fill opacity=0.3, minimum width=\inlaywidth+\inlaymargin, minimum height=\inlaywidth+\inlaymargin] at (\vx, \inlayvy )  {};
        \node[anchor=center] at (inlay_arcticsf_right.center) {\includegraphics[width=\inlaywidth]{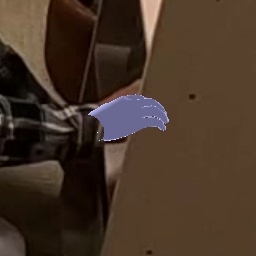}};
        \advance\vx by \colwidth
        \node (inlay_proposed_right) [inner sep=0pt, draw=none, fill=white, fill opacity=0.3, minimum width=\inlaywidth+\inlaymargin, minimum height=\inlaywidth+\inlaymargin] at (\vx, \inlayvy )  {};
        \node[anchor=center] at (inlay_proposed_right.center) {\includegraphics[width=\inlaywidth]{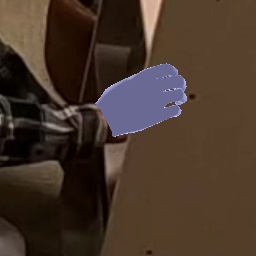}};
        \advance\vx by \colwidth
        \node (inlay_gt_right) [inner sep=0pt, draw=none, fill=white, fill opacity=0.3, minimum width=\inlaywidth+\inlaymargin, minimum height=\inlaywidth+\inlaymargin] at (\vx, \inlayvy )  {};
        \node[anchor=center] at (inlay_gt_right.center) {\includegraphics[width=\inlaywidth]{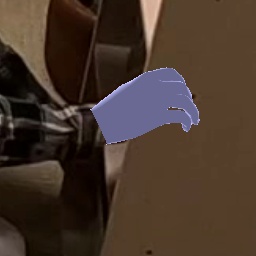}};

        \vx = \colwidth
        \advance \vx by -8
        \inlayvy = \vy
        \advance \inlayvy by -7
        \node (inlay_hamer_left) [inner sep=0pt, draw=none, fill=white, fill opacity=0.3, minimum width=\inlaywidth+\inlaymargin, minimum height=\inlaywidth+\inlaymargin] at (\vx, \inlayvy )  {};
        \node[anchor=center] at (inlay_hamer_left.center) {\includegraphics[width=\inlaywidth]{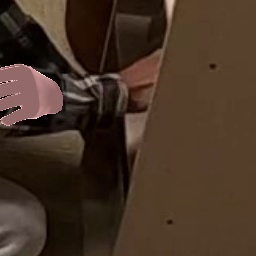}};
        \advance\vx by \colwidth
        \node (inlay_wilor_left) [inner sep=0pt, draw=none, fill=white, fill opacity=0.3, minimum width=\inlaywidth+\inlaymargin, minimum height=\inlaywidth+\inlaymargin] at (\vx, \inlayvy )  {};
        \node[anchor=center] at (inlay_wilor_left.center) {\includegraphics[width=\inlaywidth]{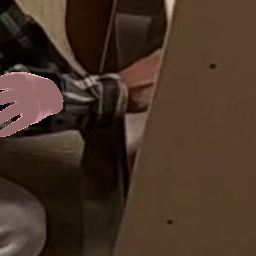}};
        \advance\vx by \colwidth
        \node (inlay_arcticsf_left) [inner sep=0pt, draw=none, fill=white, fill opacity=0.3, minimum width=\inlaywidth+\inlaymargin, minimum height=\inlaywidth+\inlaymargin] at (\vx, \inlayvy )  {};
        \node[anchor=center] at (inlay_arcticsf_left.center) {\includegraphics[width=\inlaywidth]{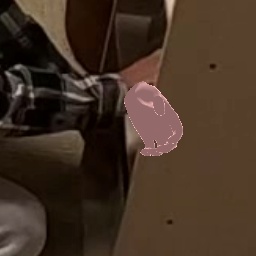}};
        \advance\vx by \colwidth
        \node (inlay_proposed_left) [inner sep=0pt, draw=none, fill=white, fill opacity=0.3, minimum width=\inlaywidth+\inlaymargin, minimum height=\inlaywidth+\inlaymargin] at (\vx, \inlayvy )  {};
        \node[anchor=center] at (inlay_proposed_left.center) {\includegraphics[width=\inlaywidth]{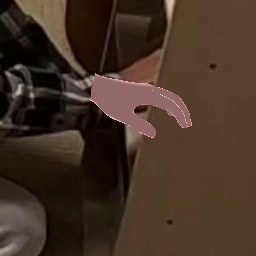}};
        \advance\vx by \colwidth
        \node (inlay_gt_left) [inner sep=0pt, draw=none, fill=white, fill opacity=0.3, minimum width=\inlaywidth+\inlaymargin, minimum height=\inlaywidth+\inlaymargin] at (\vx, \inlayvy )  {};
        \node[anchor=center] at (inlay_gt_left.center) {\includegraphics[width=\inlaywidth]{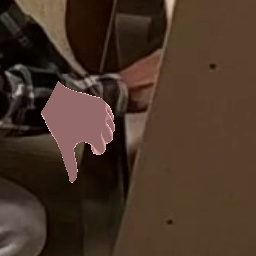}};

        \advance\vy by -\rowheight
        \node[anchor=north] at (\inputcol*\colwidth,    \vy) {\includegraphics[width=\imagewidth]{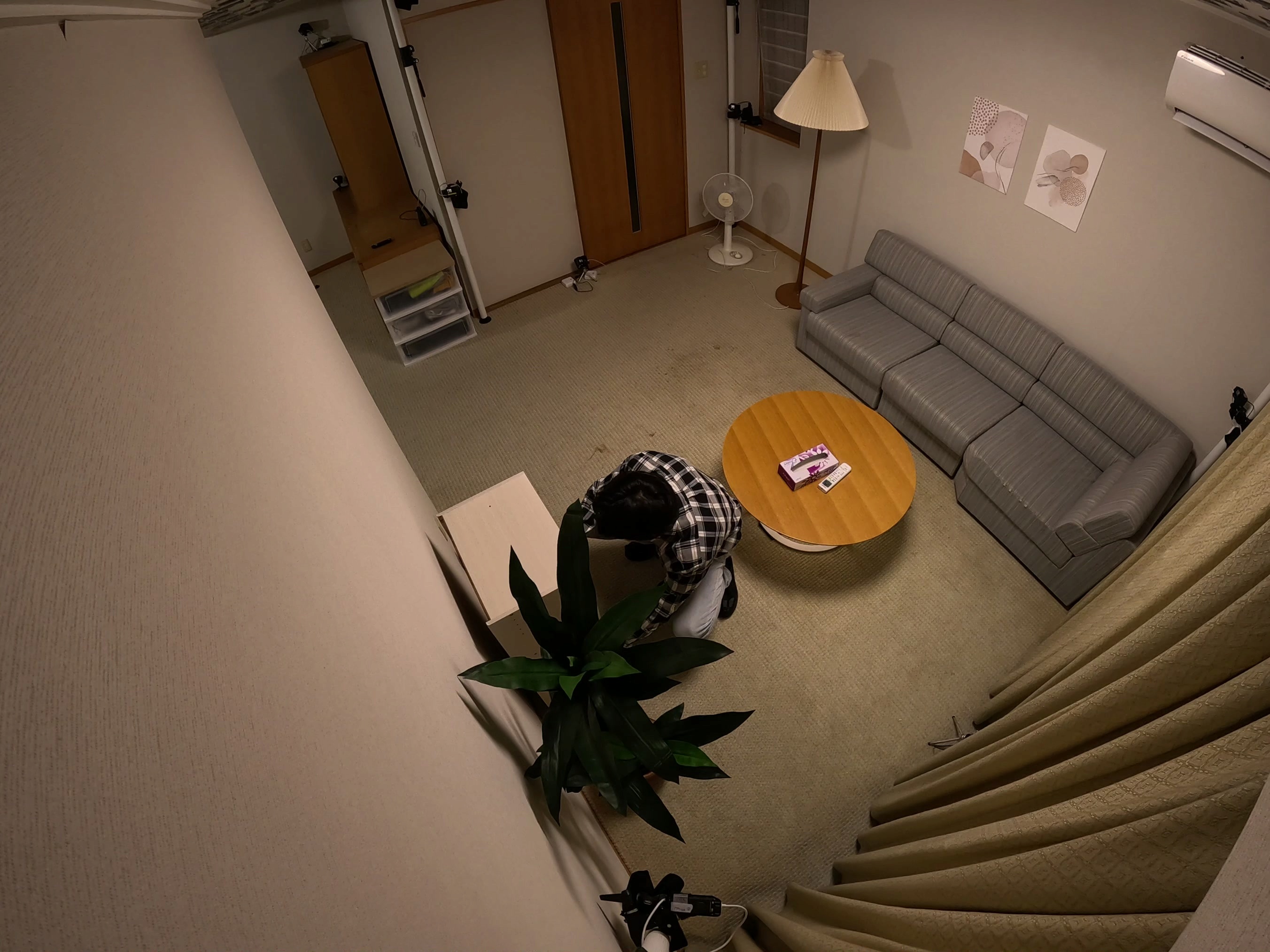}};
        \node[anchor=north] at (\proposedcol*\colwidth, \vy) {\includegraphics[width=\imagewidth]{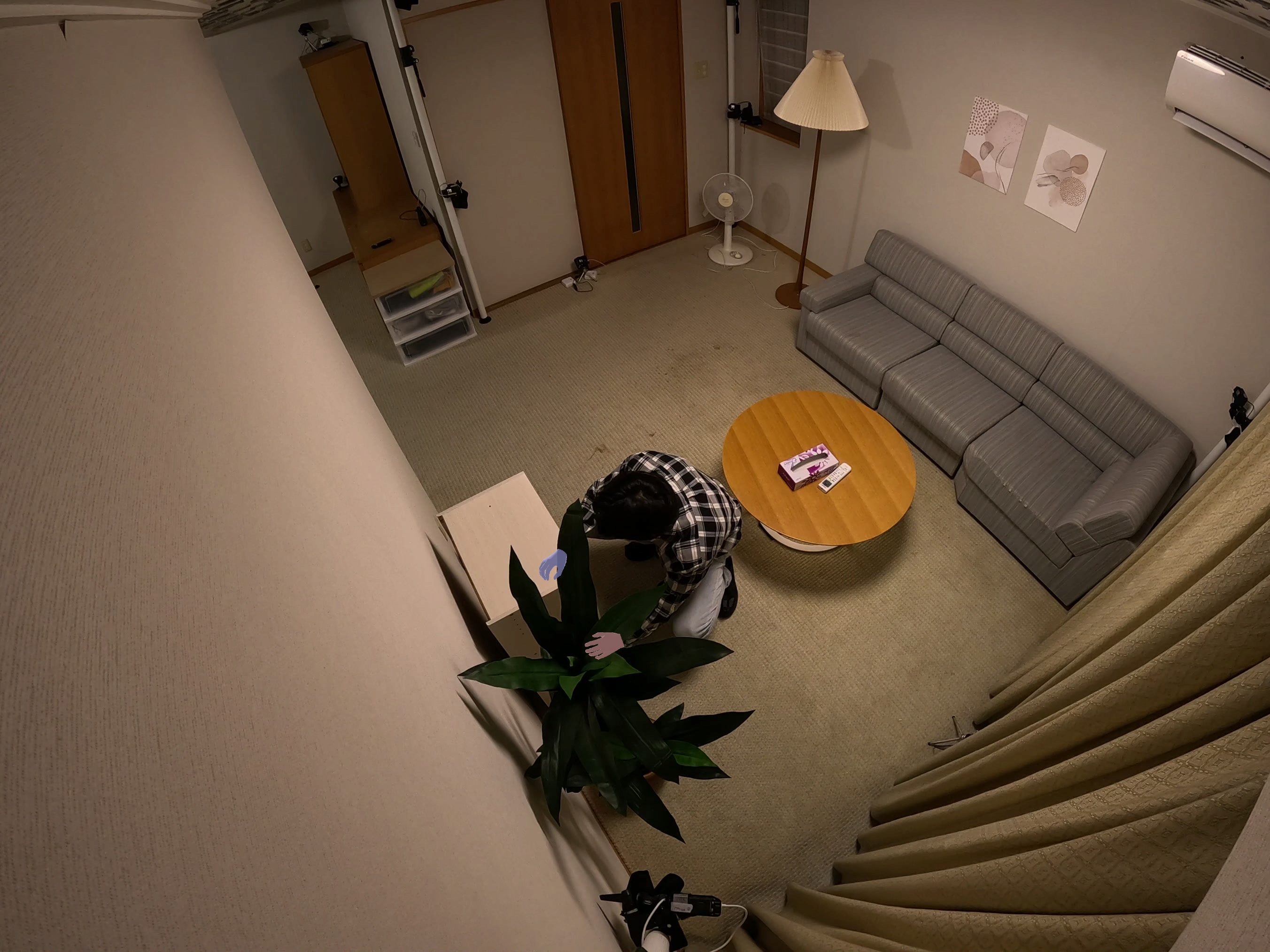}};
        \node[anchor=north] at (\hamercol*\colwidth,    \vy) {\includegraphics[width=\imagewidth]{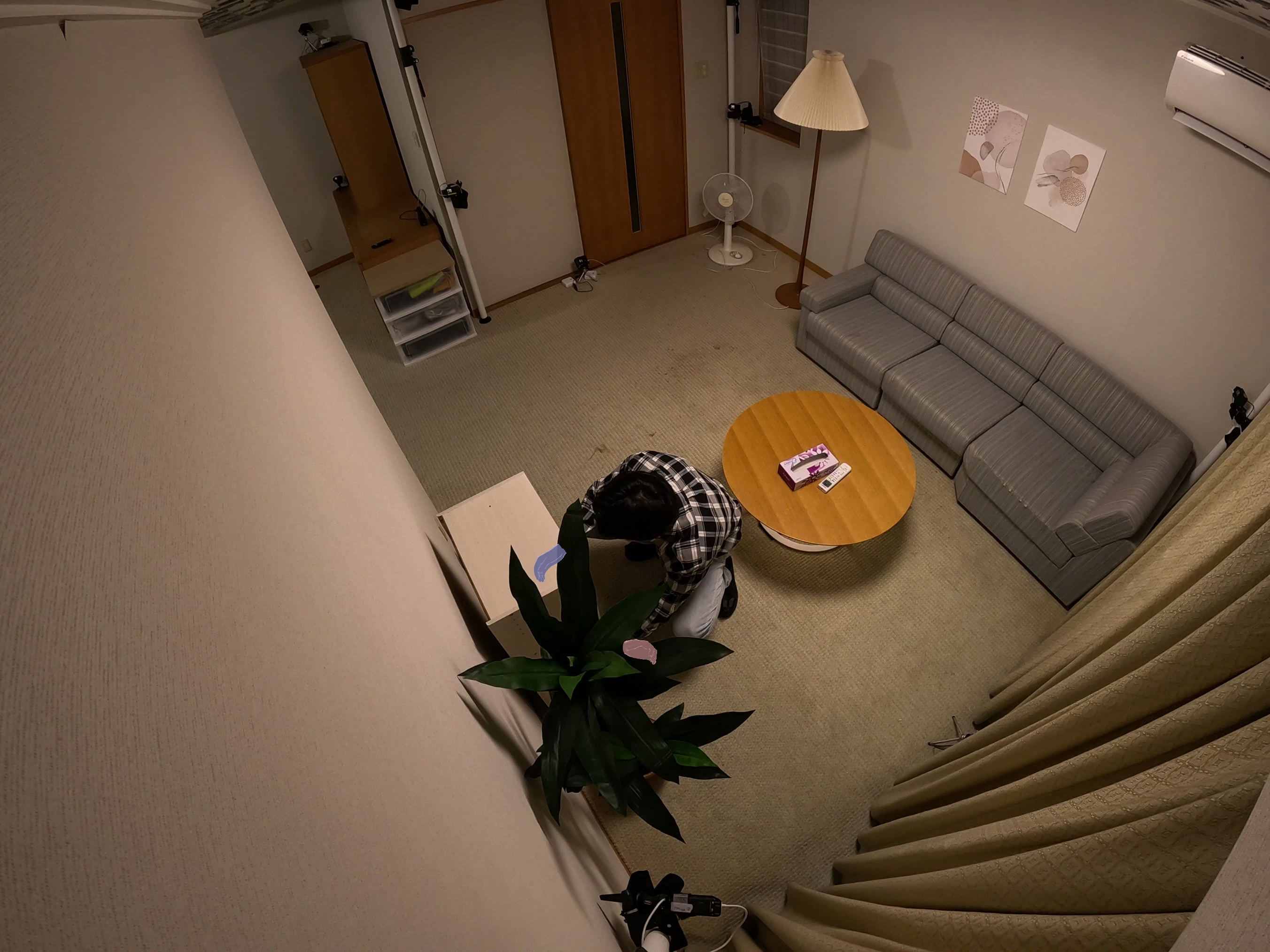}};
        \node[anchor=north] at (\wilorcol*\colwidth,    \vy) {\includegraphics[width=\imagewidth]{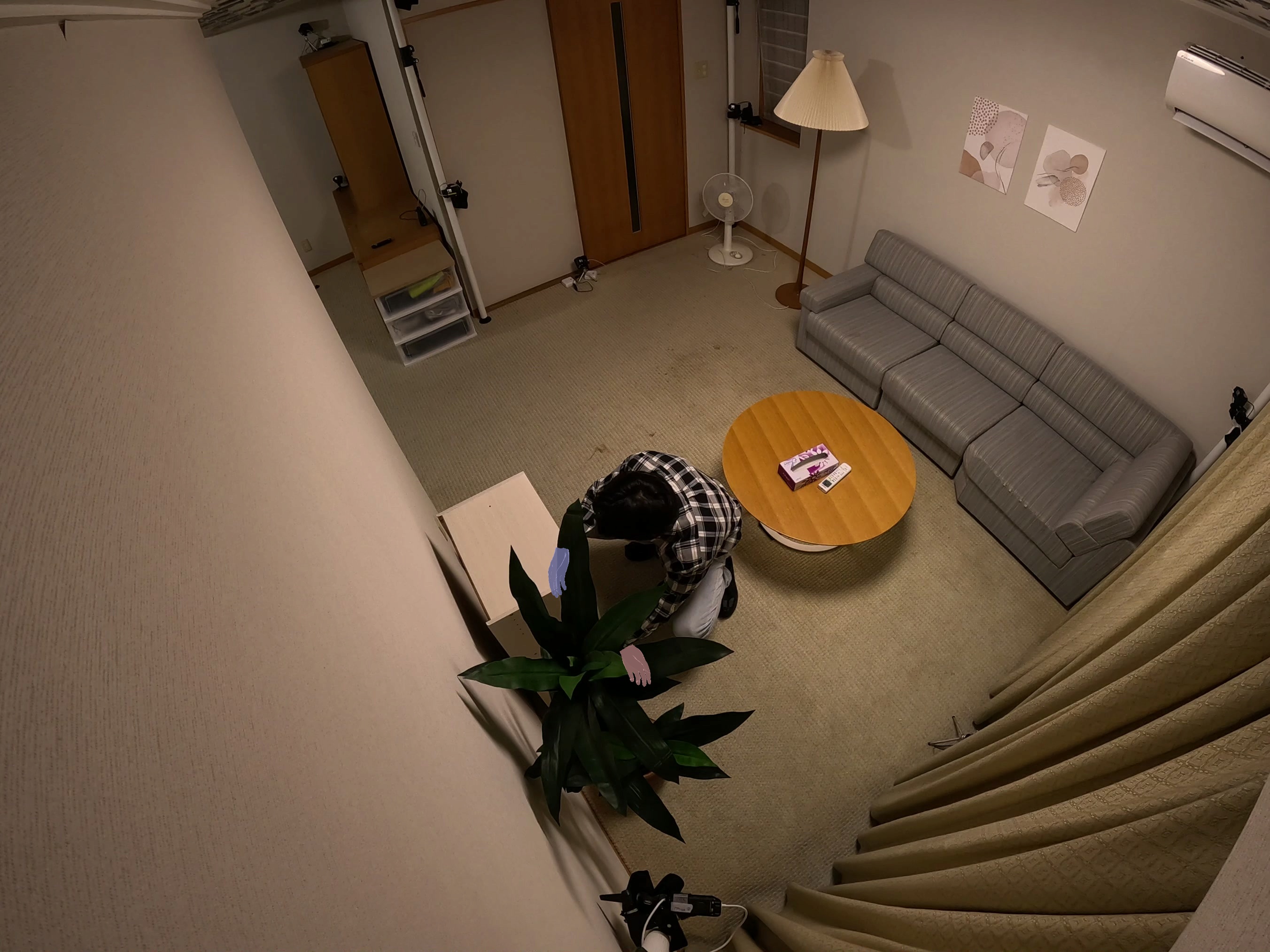}};
        \node[anchor=north] at (\arcticsfcol*\colwidth, \vy) {\includegraphics[width=\imagewidth]{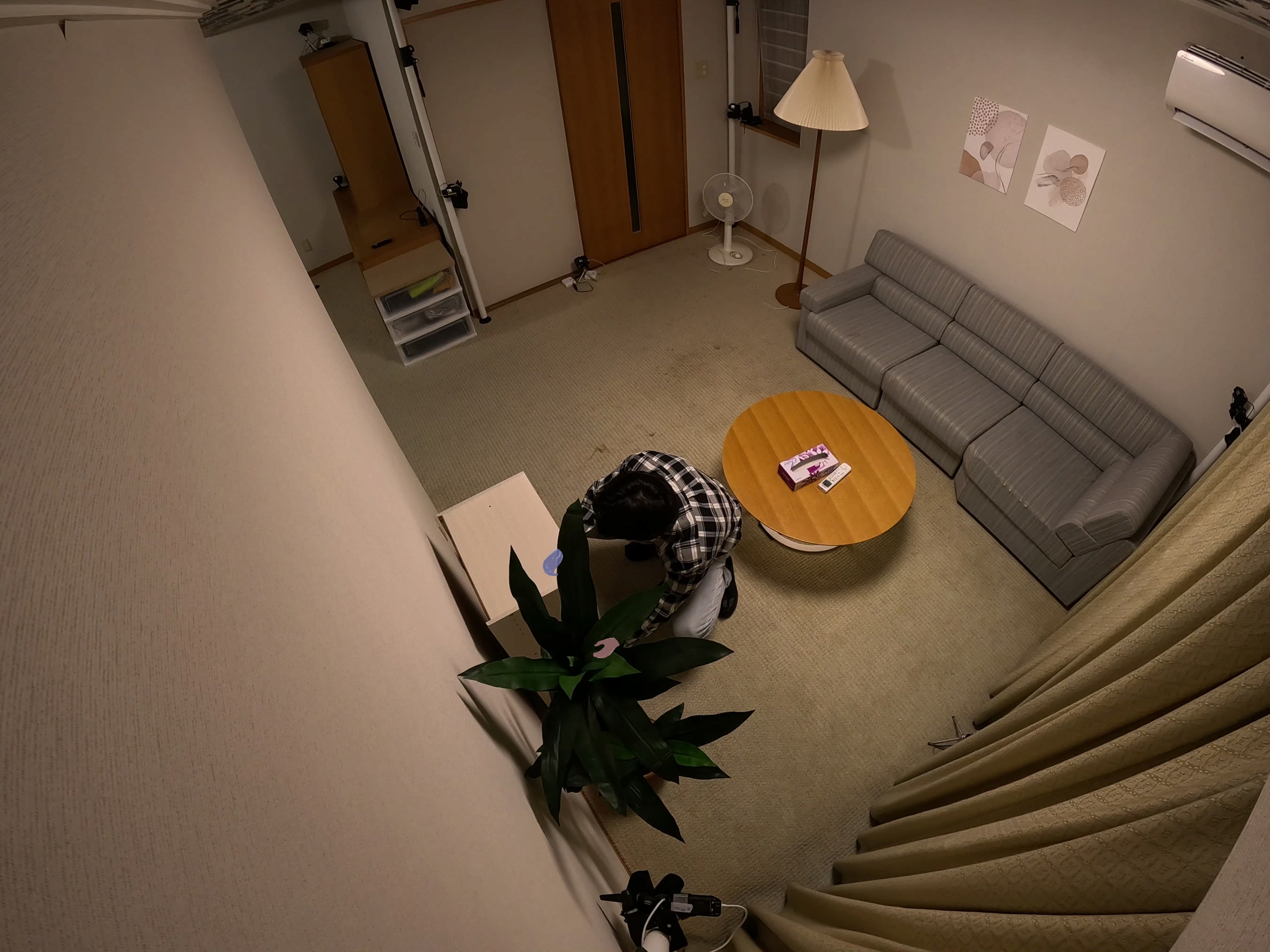}};
        \node[anchor=north] at (\gtcol*\colwidth,       \vy) {\includegraphics[width=\imagewidth]{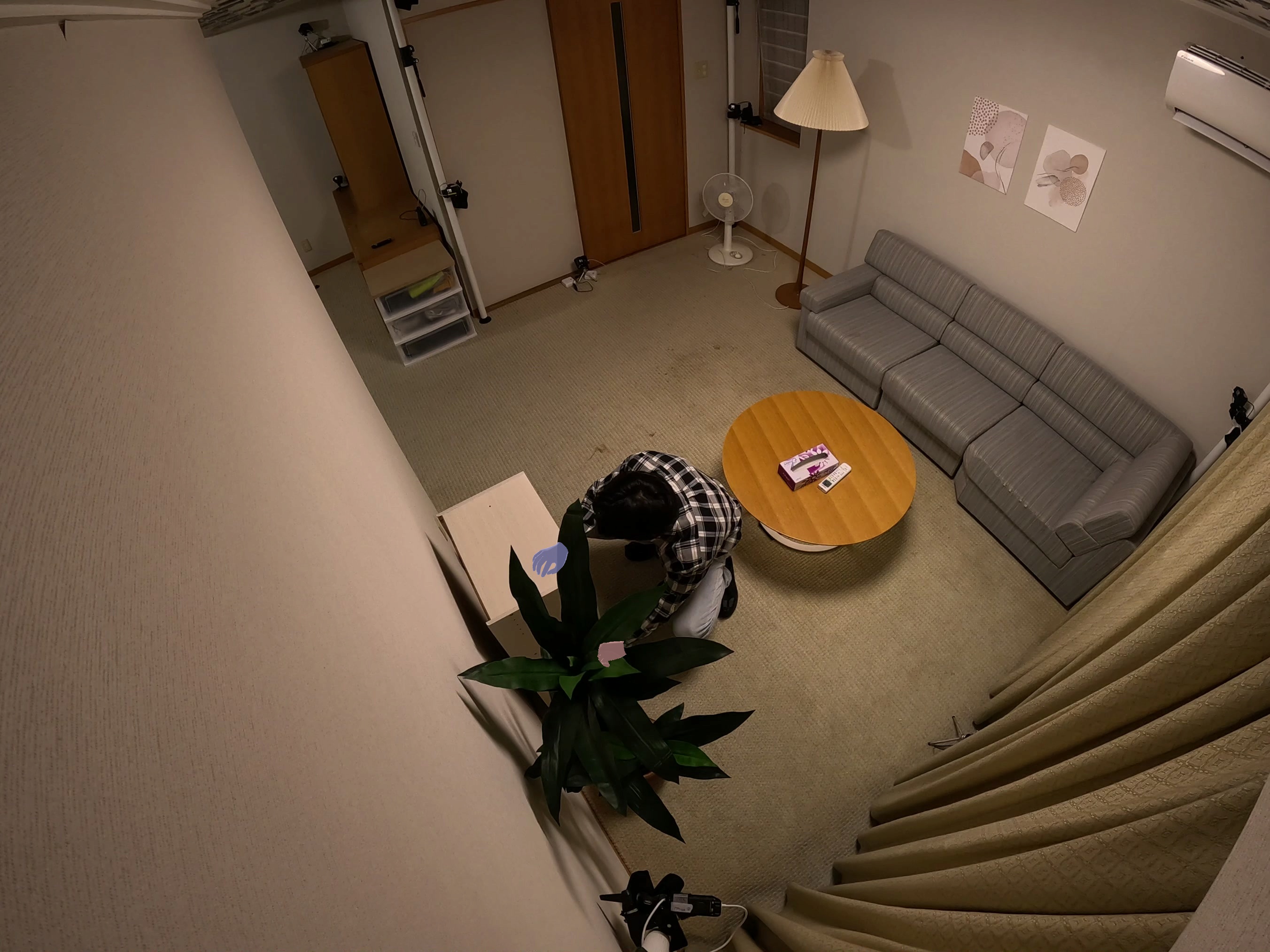}};

        \vx = \colwidth
        \advance \vx by -8
        \inlayvy = \vy
        \advance \inlayvy by -7
        \node (inlay_hamer_right) [inner sep=0pt, draw=none, fill=white, fill opacity=0.3, minimum width=\inlaywidth+\inlaymargin, minimum height=\inlaywidth+\inlaymargin] at (\vx, \inlayvy )  {};
        \node[anchor=center] at (inlay_hamer_right.center) {\includegraphics[width=\inlaywidth]{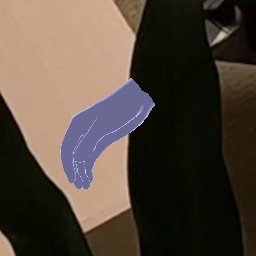}};
        \advance\vx by \colwidth
        \node (inlay_wilor_right) [inner sep=0pt, draw=none, fill=white, fill opacity=0.3, minimum width=\inlaywidth+\inlaymargin, minimum height=\inlaywidth+\inlaymargin] at (\vx, \inlayvy )  {};
        \node[anchor=center] at (inlay_wilor_right.center) {\includegraphics[width=\inlaywidth]{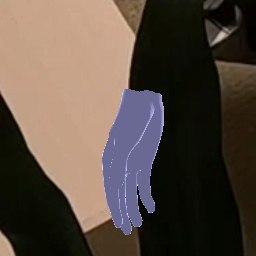}};
        \advance\vx by \colwidth
        \node (inlay_arcticsf_right) [inner sep=0pt, draw=none, fill=white, fill opacity=0.3, minimum width=\inlaywidth+\inlaymargin, minimum height=\inlaywidth+\inlaymargin] at (\vx, \inlayvy )  {};
        \node[anchor=center] at (inlay_arcticsf_right.center) {\includegraphics[width=\inlaywidth]{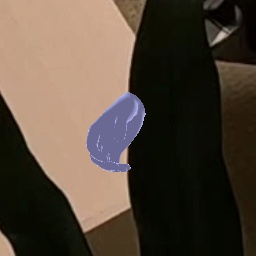}};
        \advance\vx by \colwidth
        \node (inlay_proposed_right) [inner sep=0pt, draw=none, fill=white, fill opacity=0.3, minimum width=\inlaywidth+\inlaymargin, minimum height=\inlaywidth+\inlaymargin] at (\vx, \inlayvy )  {};
        \node[anchor=center] at (inlay_proposed_right.center) {\includegraphics[width=\inlaywidth]{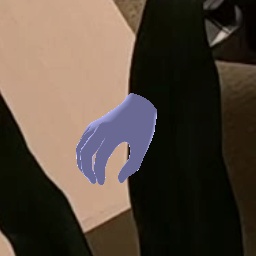}};
        \advance\vx by \colwidth
        \node (inlay_gt_right) [inner sep=0pt, draw=none, fill=white, fill opacity=0.3, minimum width=\inlaywidth+\inlaymargin, minimum height=\inlaywidth+\inlaymargin] at (\vx, \inlayvy )  {};
        \node[anchor=center] at (inlay_gt_right.center) {\includegraphics[width=\inlaywidth]{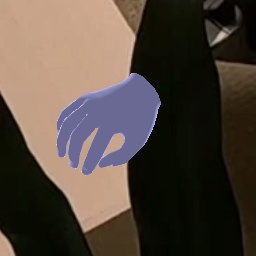}};

        \vx = \colwidth
        \advance \vx by 8
        \inlayvy = \vy
        \advance \inlayvy by -16
        \node (inlay_hamer_left) [inner sep=0pt, draw=none, fill=white, fill opacity=0.3, minimum width=\inlaywidth+\inlaymargin, minimum height=\inlaywidth+\inlaymargin] at (\vx, \inlayvy )  {};
        \node[anchor=center] at (inlay_hamer_left.center) {\includegraphics[width=\inlaywidth]{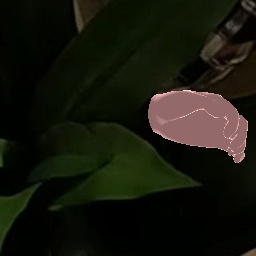}};
        \advance\vx by \colwidth
        \node (inlay_wilor_left) [inner sep=0pt, draw=none, fill=white, fill opacity=0.3, minimum width=\inlaywidth+\inlaymargin, minimum height=\inlaywidth+\inlaymargin] at (\vx, \inlayvy )  {};
        \node[anchor=center] at (inlay_wilor_left.center) {\includegraphics[width=\inlaywidth]{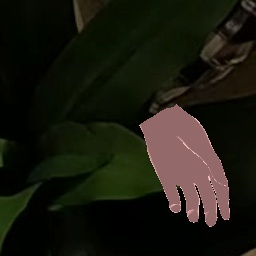}};
        \advance\vx by \colwidth
        \node (inlay_arcticsf_left) [inner sep=0pt, draw=none, fill=white, fill opacity=0.3, minimum width=\inlaywidth+\inlaymargin, minimum height=\inlaywidth+\inlaymargin] at (\vx, \inlayvy )  {};
        \node[anchor=center] at (inlay_arcticsf_left.center) {\includegraphics[width=\inlaywidth]{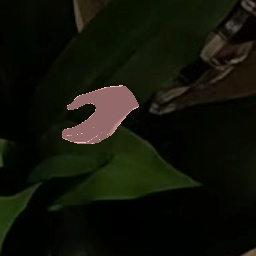}};
        \advance\vx by \colwidth
        \node (inlay_proposed_left) [inner sep=0pt, draw=none, fill=white, fill opacity=0.3, minimum width=\inlaywidth+\inlaymargin, minimum height=\inlaywidth+\inlaymargin] at (\vx, \inlayvy )  {};
        \node[anchor=center] at (inlay_proposed_left.center) {\includegraphics[width=\inlaywidth]{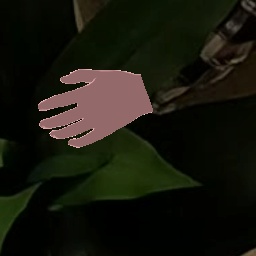}};
        \advance\vx by \colwidth
        \node (inlay_gt_left) [inner sep=0pt, draw=none, fill=white, fill opacity=0.3, minimum width=\inlaywidth+\inlaymargin, minimum height=\inlaywidth+\inlaymargin] at (\vx, \inlayvy )  {};
        \node[anchor=center] at (inlay_gt_left.center) {\includegraphics[width=\inlaywidth]{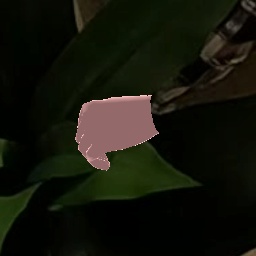}};

    \end{tikzpicture}
    \caption{Estimated 3D hand poses. The leftmost column shows the two input views and other columns show the estimates by each method, in red for the left hand and in blue for the right hand. Our method recovers 3D hand pose more accurately than the baseline methods, particularly when the hands are occluded by objects or the body.}
    \label{fig:qualitative_results}
    \vspace{-1\baselineskip}
\end{figure*}

\Cref{fig:qualitative_results} shows qualitative results of \METHODNAME{} and the baseline methods.
\METHODNAME{} estimates the hand pose accurately even when the hands are occluded by objects or the body, while the baseline methods fail to estimate the hand pose in such cases.
This shows the effectiveness of leveraging multiview input and hand-body coordination.

\begin{figure}[t]
    \centering
    \includegraphics[width=\linewidth]{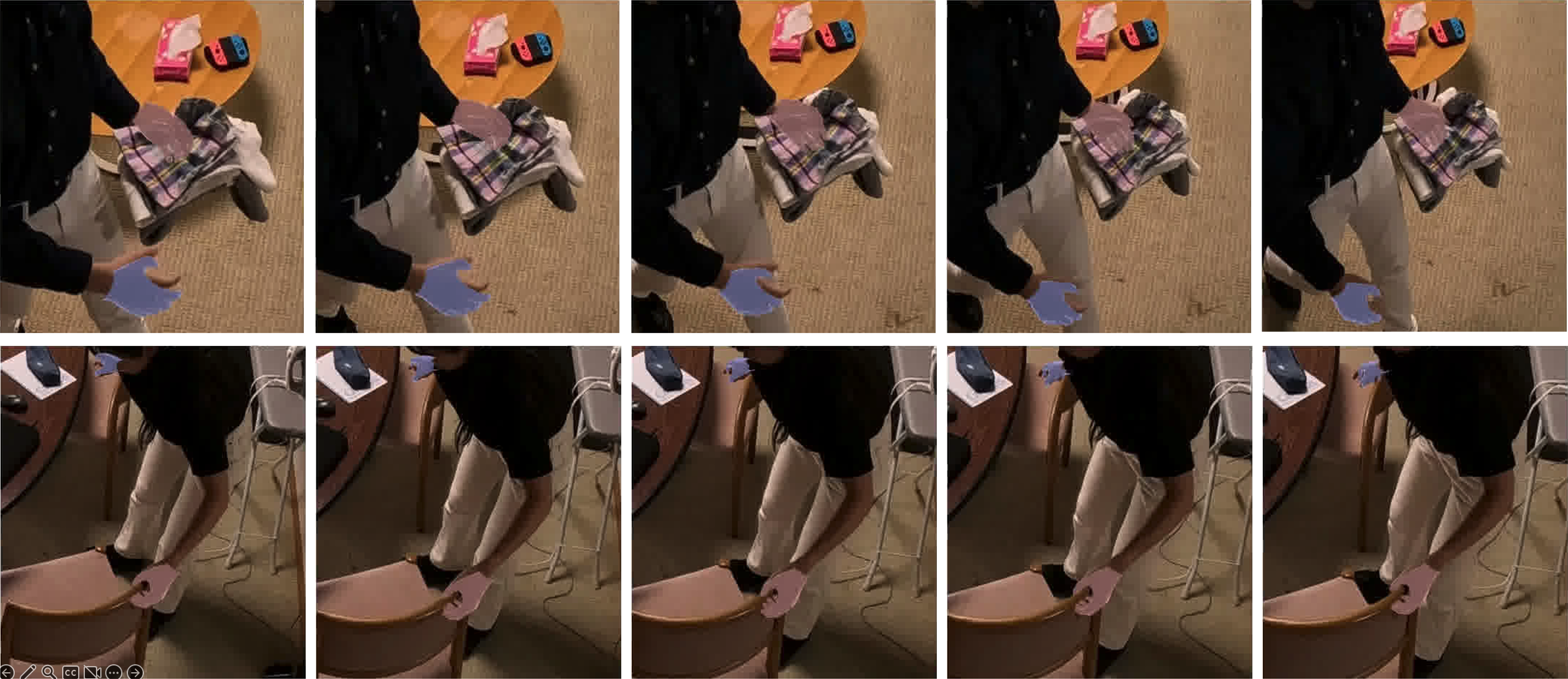}
    \caption{Trajectory of hand keypoints estimated by \METHODNAME{}. \METHODNAME{} estimates smooth and accurate 3D hand pose trajectories albeit frequent occlusions.}
    \vspace{-1\baselineskip}
    \label{fig:hand_trajectory}
\end{figure}

\Cref{fig:hand_trajectory} shows the estimated trajectories of hand poses.
\METHODNAME{} can estimate smooth and accurate hand pose trajectories even when the hands are occluded for some time thanks to its autoregressive construction.

\subsection{Effect of Hand Distance from the Camera}
To verify the accuracy of our method as a function of the distance of the participant from the camera, we separated the test set into three categories: Near, Medium, and Distant, \ie, the distance less than \qty{4}{m}, \qty{4}{m} to \qty{8}{m}, and more than \qty{8}{m}.
Since all cameras capture at the same resolution and FoV, hands that are \qty{4}{m} and \qty{8}{m} away appear \num{40} px and \num{20} px wide, respectively.
\Cref{tab:distance_comparison} shows \METHODNAME{} outperforms all the baselines in all distance ranges.
The accuracy of \METHODNAME{} in the Distant range is comparable to that of the best baseline method in the Near range, which shows the effectiveness of \METHODNAME{} for far-view 3D hand pose estimation.

\begin{table}[t]
    \centering
    \SetTblrInner{rowsep=2pt,colsep=6pt,stretch=1,abovesep=2pt}
    \begin{tblr}{@{}lrrr@{}}
        \toprule
                                            & \SetCell[c=3]{c} Distance \\
        \cmidrule[lr]{2-4}
        Method                              & Near                       & Medium         & Distant        \\
        \midrule
        HaMeR~\cite{hamer}                  & 14.90                      & 15.87          & 16.22          \\
        WiLoR~\cite{wilor2024}              & 15.36                      & 16.15          & 16.32          \\
        ArcticNet-SF~\cite{fan2023arctic}   & 13.92                      & 14.40          & 15.48          \\
        ArcticNet-LSTM~\cite{fan2023arctic} & 14.60                      & 14.93          & 15.40          \\
        \midrule
        Body2Hands~\cite{body2hands2021}    & 15.92                      & 16.56          & 17.47          \\
        \midrule
        \METHODNAME{} (Ours)                & \textbf{12.96}             & \textbf{13.44} & \textbf{14.86} \\
        \bottomrule
    \end{tblr}
    \caption{Comparison of 3D hand pose estimation accuracy (PA-MPJPE, mm) for different ranges of distance. Our method outperforms all the baselines in all distances.}
    \label{tab:distance_comparison}
    \vspace{-1\baselineskip}
\end{table}

\subsection{Generalization to Different Environments}
\METHODNAME{} assumes known relative camera poses, but its inference is invariant to them as it embeds the ray directions in the first camera's coordinate system.
We verified this by training \METHODNAME{} on data from only one of the two rooms in our dataset and testing it on the other. Our method achieves a PA-MPJPE of \qty{15.12}{mm} and a joint angle error of \ang{14.93}, which is comparable to the performance when the training and test data are from the same room.

We also evaluate \METHODNAME{} on the ARCTIC dataset~\cite{fan2023arctic} to verify its generalization capability to different datasets. Our method achieved an MPJPE of \qty{26.4}{mm} without fine-tuning, comparable to ArcticNet-LSTM~\cite{fan2023arctic}, which was specifically designed for this dataset (\qty{22.9}{mm}).
This result shows that \METHODNAME{} can generalize to different camera configurations, environments, and actions.

\subsection{Ablation Studies}
We compare the accuracy of \METHODNAME{} with and without each model component.
\Cref{tab:ablation_study} shows that removing components (multiview input, body pose, autoregressive estimation, and ray embedding after the second and later frames) drops accuracy, which indicates the effectiveness of each component. All of the differences are statistically significant with Welch's one-tailed t-test (\eg, $p=0.022$ for the closest pair).
Notably, even our single-view variant achieves higher accuracy than the baselines, which highlights the importance of leveraging hand-body coordination.

\begin{table}[t]
    \centering
    \SetTblrInner{rowsep=2pt,colsep=6pt,stretch=1,abovesep=2pt}
    \begin{tblr}{@{}lrr@{}}
        \toprule
        Variant            & PA-MPJPE (mm)            & Joint Angles (deg)        \\
        \midrule
        w/o multiview      & \num{15.12(439)}         & \num{15.60(596)}         \\
        w/o body pose      & \num{14.64(452)}         & \num{14.64(542)}         \\
        w/o autoregressive & \num{15.06(458)}         & \num{16.24(555)}         \\
        w/o ray embd.      & \num{14.75(474)}         & \num{14.95(544)}         \\
        \midrule
        Proposed           & $\mathbf{14.47}\pm 4.90$ & $\mathbf{14.48}\pm 5.34$ \\
        \bottomrule
    \end{tblr}
    \caption{Ablation studies of \METHODNAME{}. We compare accuracy with and without each key component (multiview input, body pose, autoregressive estimation, and ray embedding) using the mean and standard deviation of errors. Removing any of these components degrades accuracy, demonstrating the effectiveness of each component.}
    \label{tab:ablation_study}
    \vspace{-1\baselineskip}
\end{table}

\section{Conclusions}
In this paper, we introduced \METHODNAME, a novel method for 3D hand pose estimation from afar.
\METHODNAME leverages multiview observations of hand-body coordination and their temporal variation to robustly estimate 3D hand pose when the hands are not clearly visible, \ie, at low resolution and partially to fully occluded, due to the large distance from static cameras installed at room corners. We also introduced \DATASETNAME, a first-of-its-kind large-scale dataset that captures natural daily human activities in real rooms from wall-mounted cameras, fully annotated with accurate 3D hand pose and position.
We hope that these two contributions will serve as a sound foundation for future research on understanding human behavior from passive observations in daily life.

\section{Acknowledgements}
This work was in part supported by JSPS KAKENHI 21H04893, JST JPMJAP2305, and RIKEN GRP.

\appendix

\section{Dataset Capture Setup}

\begin{figure}[h]
    \centering
    \includegraphics[width=\linewidth]{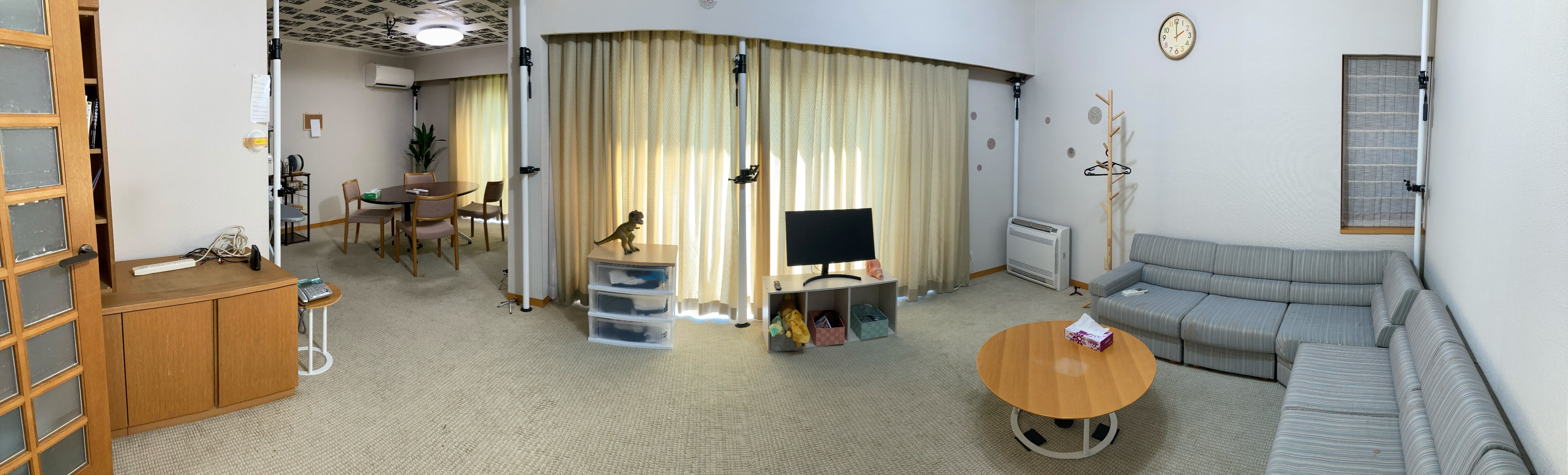}
    \caption{The capture setup of \DATASETNAME{} dataset. The environment is a real living room with various pieces of furniture and everyday objects. Note that there are no visible markers except for those on the ceiling, which are barely visible from the fixed-view cameras mounted at the ceiling corners. The participant activities are captured without any obstruction, visible markers, or cameras.}
  \label{fig:environment_overview}
\end{figure}

\Cref{fig:environment_overview} shows the overview of the capture environment.
\Cref{fig:camera_arrangement} shows the camera arrangement of the dataset. Note that the green cameras on the ground are used solely to locate the ArUco~\cite{aruco} markers on the ceiling, and their video feeds are not included in the dataset.

\definecolor{redcamera}{HTML}{ff80a6}
\definecolor{bluecamera}{HTML}{80e5ff}
\definecolor{greencamera}{HTML}{8eab36}
\begin{figure}[h]
    \centering
    \includegraphics[width=\linewidth]{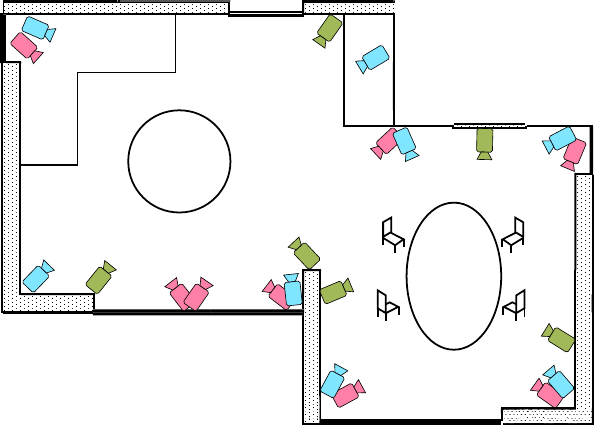}
    \caption{Camera arrangement of the \DATASETNAME{} dataset. 
        \textcolor{bluecamera}{Blue} cameras are installed on the ceiling, while \textcolor{redcamera}{red} cameras are on the walls (from the height of the chest to the hip).
        \textcolor{greencamera}{Green} cameras are auxiliary cameras that capture ceiling markers for estimating the chest camera's location.
    }
  \label{fig:camera_arrangement}
\end{figure}

\section{Camera Pose Estimation by Markers and SLAM}
As explained in the main paper, we combine two methods to estimate the chest camera trajectory in the world coordinate system.
The first method is to use ArUco~\cite{aruco} markers placed on the ceiling of the room.
The second method is to use ORB-SLAM3~\cite{orb-slam3} to obtain dense camera trajectories.
For each frame in the video $i \in \left\{1\ldots T\right\}$, the first method can estimate camera pose $\bm{P}^\text{M}_i = (\bm{R}^\text{M}_i, \bm{t}^\text{M}_i)$ in the world coordinate system as the marker positions are calibrated beforehand.
Pose estimation using markers, however, is not possible or can be noisy when the markers are barely visible, which often occurs when the hands occlude them or when the subject is bending down.
We therefore estimate chest camera trajectory $\bm{P}^\text{S}_i = (\bm{R}^\text{S}_i, \bm{t}^\text{S}_i)$ using ORB-SLAM3~\cite{orb-slam3} too, which is robust against temporary occlusion. Finally, we align the two camera trajectories by solving the following equation through Procrustes analysis
\begin{equation}
    \argmin_{\bm{R},\bm{t},s} \sum_{i=1}^{T} \|\bm{t}^\text{M}_i - s\left(\bm{R}\bm{t}^\text{S}_i+\bm{t}\right)\|^2_2\,.
\end{equation}
We then use $\bm{P}_i = \left(\bm{R} \bm{R}^\text{S}_i, s\left(\bm{R}\bm{t}^\text{S}_i+\bm{t}\right)\right)$ as the chest camera trajectories.

\section{Wrist Triangulation from Wall Cameras}
We first detect the subject's body keypoints using an existing body pose estimation model~\cite{vitpose}.
For robust 3D body keypoint recovery from noisy 2D body pose estimates, we adaptively select ``reliable'' cameras for each keypoint and triangulate only from their observations.
For each camera pair, we triangulate the body keypoints using the 2D keypoint estimates. If the 3D triangulation error, defined as the distance between the triangulated 3D keypoint and the rays from the two cameras to the detected 2D keypoints, is below a certain threshold, we consider the cameras in the pair reliable.
Using the reliable cameras, we triangulate the body keypoints and recalculate the 3D triangulation error. If the triangulation error falls below a certain threshold, we consider the triangulation successful.

We then define the ``hand keypoint'' from the body keypoints, computed as an interior point between the 3D elbow and wrist keypoints. We determine the division ratio that minimizes the rendering error of the hand mesh whose wrist is located at the hand keypoint calculated from the ratio. To do this, we estimate the hand region using SAM2~\cite{sam2}, compare it with the hand mesh rendered on the fixed-camera images, and choose the division ratio that maximizes their intersection.

\section{Details of \METHODNAME{} and Experimental Setup}

\paragraph{Model Parameters}
For the Pose Encoder, we use a lightweight Transformer encoder with 6 layers, each layer having 8 attention heads and a hidden dimension of 256.
Therefore, input embedding $e_n^t$ has a dimension of $2304 = 256 + 1024 + 1024$.
For the Transformer network of the \METHODNAME{}, we use a 6-layer encoder-decoder architecture with 4 attention heads with a hidden dimension of 1024.

\paragraph{Experimental Setup}
During training, we use the AdamW optimizer~\cite{AdamW} with a learning rate of \num{2e-5} and a weight decay of \num{1e-4}.
For the loss function, we set $\lambda_{R} = \lambda_{T} = \lambda_{\beta} = \lambda_{\theta}= \lambda_{J} = 1 $ and $ \lambda_\text{C} = 0.1$.
We set the temporal batch size (the number of frames to estimate autoregressively) to \num{4} and use 8 such temporal batches per training batch. Temporal stride is set to \num{4}, meaning that the input frames are sampled at \qty{15}{FPS}.
The participants are randomly split into training (40 subjects), validation (5 subjects), and test (5 subjects) sets.
We run the training for 60k iterations for the first and second stages, respectively.
The training was run on an NVIDIA H200 GPU and took 24 hours per stage.
Inference of the model runs at 16.3 FPS on an NVIDIA A6000 GPU.
{\small
\bibliographystyle{ieee}
\bibliography{main}
}

\end{document}